\documentclass{article}

\PassOptionsToPackage{numbers, compress}{natbib}

\usepackage[preprint]{neurips_2025}

\usepackage{hyperref}
\usepackage{mathtools}
\usepackage{xcolor}
\usepackage{amsmath}
\usepackage{amssymb}
\usepackage{subcaption}
\usepackage{multirow}
\usepackage{booktabs}
\usepackage{wrapfig}
\usepackage[ruled,vlined]{algorithm2e}
\usepackage{pifont}
\usepackage{url}      
\usepackage[symbol]{footmisc}
\renewcommand{\thefootnote}{\fnsymbol{footnote}}

\newcommand{\cmark}{\ding{51}}
\newcommand{\xmark}{\ding{55}}

\newcommand\myeq{\stackrel{\text{\tiny def}}{=}}

\newcommand{\fraci}[2]{{#1}/{#2}}
\newcommand{\apg}{\ensuremath{\tilde{\nabla}\theta_i^t}}
\newcommand{\apgnot}{\ensuremath{\tilde{\nabla}\theta_i}}
\newcommand{\apgnotnoi}{\ensuremath{\tilde{\nabla}\theta}}
\newcommand{\cumapg}{\ensuremath{\tilde{\nabla}\theta^\text{cum}}}
\newcommand{\apgl}{\ensuremath{\tilde{\nabla}\theta_{l,i}^t}}

\hypersetup{
colorlinks=true,
linkcolor=black,
citecolor=black,
urlcolor=black
}

\title{A Scalable Hybrid Training Approach for Recurrent Spiking Neural Networks}

\author{%
  Maximilian Baronig\thanks{These authors contributed equally.} \\
  \texttt{baronig@tugraz.at} \\
  \And
  Yeganeh Bahariasl\footnotemark[1] \\
  \texttt{yeganeh.bahariasl@tugraz.at} \\
    \And
  Ozan Özdenizci \\
  \texttt{oezdenizci@tugraz.at} \\
      \And
  Robert Legenstein \\
  \texttt{robert.legenstein@tugraz.at} \\ \\
  Institute of Machine Learning and Neural Computation, Graz University of Technology, Austria
}
\begin{document}
\maketitle
\begin{abstract}
Recurrent spiking neural networks (RSNNs) can be implemented very efficiently in neuromorphic systems. Nevertheless, training of these models with powerful gradient-based learning algorithms is mostly performed on standard digital hardware using Backpropagation through time (BPTT). However, BPTT has substantial limitations. It does not permit online training and its memory consumption scales linearly with the number of computation steps.
In contrast, learning methods using forward propagation of gradients operate in an online manner with a memory consumption independent of the number of time steps. These methods enable SNNs to learn from continuous, infinite-length input sequences. Yet, slow execution speed on conventional hardware as well as inferior performance has hindered their widespread application. 
In this work, we introduce HYbrid PRopagation (HYPR) that combines the efficiency of parallelization with approximate online forward learning. Our algorithm yields high-throughput online learning through parallelization, paired with constant, i.e., sequence length independent, memory demands. 
HYPR enables parallelization of parameter update computation over the sub sequences for RSNNs consisting of almost arbitrary non-linear spiking neuron models. 
We apply HYPR to networks of spiking neurons with oscillatory subthreshold dynamics. We find that this type of neuron model is particularly well trainable by HYPR, resulting in an unprecedentedly low task performance gap between approximate forward gradient learning and BPTT.
\end{abstract}
\section{Introduction}
Spiking neural networks (SNNs) \citep{maass1997networks} and in particular recurrent SNNs (RSNNs) constitute the basis of energy-efficient computation in the brain \citep{gerstner2014neuronal} and in neuromorphic hardware \citep{young2019review}. While RSNNs can be implemented efficiently in neuromorphic systems, training of these models with powerful gradient-based learning algorithms is mostly performed on standard digital hardware such as graphical processing units (GPUs) using BPTT --- the gold standard training method for spiking and non-spiking recurrent neural networks. 

However, BPTT requires an expensive backward pass through the entire sequential computation process, with a memory and time complexity scaling linearly with the number of computation steps. The sequential nature of this algorithm introduces a computational bottleneck on GPUs, where the unrolled computational graph needs to be processed state-by-state in the backward pass, hindering parallelization. In artificial neural network models for sequence processing, this bottleneck has recently been addressed by parallelizable models \citep{guCombiningRecurrentConvolutional2021,guHiPPORecurrentMemory2020,guptaDiagonalStateSpaces2022,orvietoResurrectingRecurrentNeural2023}, achieving significantly increased throughput via more exhaustive utilization of GPU resources. The parallelization of these sequence models, which are often referred to as deep state-space models (SSMs), is achieved by removing the non-linearity between RNN state transitions. 
In the realm of SNNs, corresponding parallelizable SNN models have been proposed \citep{liParallelSpikingUnit2024,huangPRFParallelResonate2024,xueChannelwiseParallelizableSpiking2025,bal2025pspikessm}. However, in order to utilize the parallel processing power of GPUs, significant departures from the fundamental properties of spiking neuron models have to be accepted. First, nonlinear behavior of neural dynamics such as the membrane potential reset after a spike have to be avoided, and second, no recurrent connections are possible within the network. 
In addition, the fundamental limitations of BPTT remain: In addition to its sequence-length dependent memory consumption, BPTT can only operate in an offline manner, since it requires processing of the full input time series before the backward pass can be initiated. This principle is fundamentally incompatible with neuromorphic hardware \citep{bellecSolutionLearningDilemma2020}.

To address this issue, online methods of gradient learning have been proposed, with real-time recurrent learning (RTRL) \citep{williamsLearningAlgorithmContinually1989a} as its initial form, where instead of a separate backward pass, the gradients are accumulated during the forward pass.
While the memory requirement of RTRL is independent of the computation length, it scales as $\mathcal{O}(N^2)$ with the number of parameters $N$, which renders this and related algorithms infeasible in practice (see Appendix \ref{app:overhead}). 
To overcome this issue, approximations have been introduced that offer a trade-off between memory overhead and gradient approximation accuracy. One such approximation is e-prop (eligibility propagation) \citep{bellecSolutionLearningDilemma2020}, where the pathways of gradient flow through the recurrent synaptic connections are disregarded, while gradients through the neuron dynamics are forward propagated as in RTRL. Approximate forward propagation algorithms can be applied to virtually all reasonable spiking neuron models and recurrent network architectures. However, while they can be implemented efficiently in neuromorphic hardware \citep{frenkel}, they do not take advantage of parallelization in the time domain, which makes training on long sequential input extremely time consuming on standard hardware such as GPUs, hindering progress in this direction of research.

Here, we introduce HYbrid PRopagation (HYPR), a method that combines approximate online forward learning with segment-wise parallel approximate backpropagation to enable partial parallelization during training. By ignoring the gradient pathways through recurrent connections, we parallelize BPTT within each new sequence segment. The resulting back-propagated gradients are then combined with the forward-propagated gradients from previous segments, allowing for infinite training context length with constant memory complexity. 
We show, that the combination of parallelization and approximate forward propagation yields the best of both worlds: high-throughput segment-wise online learning through parallelization paired with constant, i.e., sequence length independent, memory demands and high accuracy through powerful neuron models. HYPR enables parallelization of parameter update computation over sequence segments for recurrent SNNs and almost arbitrary spiking and non-spiking neuron models. This holds even for neuron models that are not inherently parallelizable due to non-linearities in the state transition function, for example due to a spike-triggered reset mechanism, as it is commonly used in leaky integrate-and-fire (LIF) or adaptive LIF neurons. We show that even in a medium-sized network, this segment-wise parallelization of parameter update computations results in a $108\times$-speedup of training compared to the mathematically equivalent fully-online algorithm e-prop, when executed on a GPU. 

Recent work on RSNNs \citep{baronig2025advancingspatiotemporalprocessingspiking,bittar2022surrogate,higuchi2024balanced} suggests that oscillatory state dynamics of neural elements can be very beneficial for sequence processing. Since such neuron models introduce complex neuon state dynamics, they may be particularly well-suited for approximate forward propagation algorithms such as HYPR, which propagate gradients through the state dynamics but neglect those through network recurrencies. We therefore applied HYPR to networks consisting of spiking neurons with oscillatory subthreshold dynamics \citep{baronig2025advancingspatiotemporalprocessingspiking,higuchi2024balanced}. We found that using HYPR to train such SNNs clearly reduces the gap to BPTT-trained SNNs. 
\section{Related Work}
Several variants and approximations of forward- and back-propagation of gradients have been proposed. For example, the sequence-length dependent memory consumption of BPTT has been addressed in truncated BPTT \citep{williams1990efficient}, where gradients are only back-propagated for a predefined number of time steps, enabling constant memory complexity. An obvious consequence of this truncation however is that temporal credit assignment beyond its context window is impossible, rendering it infeasible to learn tasks where long-term dependencies occur.

The exact forward learning RTRL \citep{williamsLearningAlgorithmContinually1989a} on the other hand naturally solves the problem of supporting temporal credit assignment across infinite context length under constant memory, at the cost of very high memory demands of $\mathcal{O}(N^2)$ and slow execution time due to high memory I/O load (see Appendix \ref{app:overhead}). This complexity has been reduced successfully by approximation \citep{tallec2018unbiased,NEURIPS2018_dba132f6,Menick2021PracticalRT,Silver2022LearningBD} of the full sensitivity matrix or by discarding certain pathways of the gradients as in e-prop \citep{bellecSolutionLearningDilemma2020} or OTTT \citep{onlinetrainingthroughtime}. Interestingly, omission of certain pathways of the gradient can be applied to RTRL and BPTT alike. It reduces the computational complexity of both algorithms and, as we show in this work, enables BPTT to be efficiently parallelized over the time dimension. As discussed in \citep{schmidhuber1992fixed,irieExploringPromiseLimits2024}, combining forward gradient learning for infinite context length with segment-wise backward gradient learning for efficiency constitutes a potentially powerful and efficient hybrid of both methods. Despite this promising perspective, this research direction remains surprisingly underexplored. Our work shows how a combination of e-prop and segment-wise backward gradient accumulation can partially parallelize training.

Online learning has also been discussed in the context of parallelizable models: \cite{irieExploringPromiseLimits2024} and \cite{zucchetOnlineLearningLongrange2023} show that in parallelizable models the memory complexity of RTRL can be significantly reduced despite using exact gradients when nonlinear inter-dependencies of neurons of the same layer are removed. While being applicable to the family of parallelizable spiking networks \citep{liParallelSpikingUnit2024,huangPRFParallelResonate2024,bal2025pspikessm,fangParallelSpikingNeurons2023}, this principle is fundamentally incompatible with standard RSNNs such as vanilla recurrent LIF networks.
\section{Background}
\subsection{Nonlinear neuron models}
\begin{figure}[!t]
    \centering
    \includegraphics[width=\linewidth]{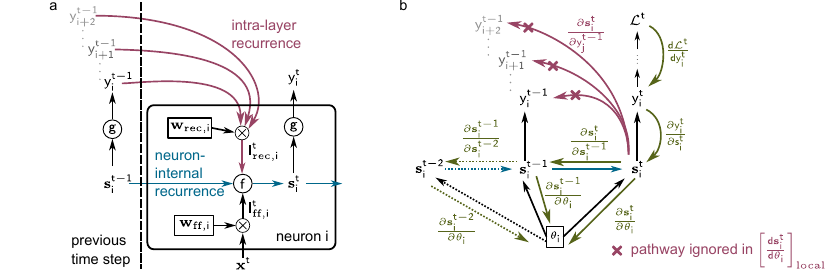}
    \caption{\textbf{a} Generic neuron model framework considered in this work. We differentiate between intra-layer neuron recurrence through explicit recurrent weighted connections (magenta) and implicit recurrence through the state-to-state transition (blue). \textbf{b} The local gradient 
    $\left[\fraci{d \mathbf{s}_i^t}{d\mathbf{\theta}_i}\right]_\text{local}$ 
    is given by the pathway through the states of neuron $i$ without considering indirect influence through outputs of other neurons. The magenta pathways are ignored in the local gradient.}
    \label{fig:neuron_model}
\end{figure}

Consider a single neuron in a layer of recurrently connected (potentially spiking) neurons with an arbitrary, possibly non-linear, state transition function $f$ and an output function $g$ given by
\begin{equation}
\label{eq:state_transition}
    \mathbf{s}_i^{t} = f(\mathbf{s}^{t-1}_i, I_i^t), \qquad \qquad     y_i^t = g(\mathbf{s}^t_i),
\end{equation}
where $\mathbf{s}^t_i \in \mathbb{R}^k$ denotes the $k$-dimensional state of neuron $i$ at time step $t$, $I_i^t \in \mathbb{R}$ the neuron input and $y_i^{t} \in \mathbb{R}$ the neuron output, as shown in Fig.~\ref{fig:neuron_model}a. The state dimension $k$ varies between different neuron models. In spiking neuron models, the output function $g$ is usually the Heaviside step function, which is not differentiable. As is common practice --- and also adopted in this work --- the partial derivative $\fraci{\partial y_i^t}{\partial \mathbf{s}_i^t}$ is hence approximated using a surrogate derivative \citep{esser_surrogate,bellec2018long} (see Appendix \ref{app:training_and_hyperparams}).
The scalar neuron input $I_i^t$ is composed of a feed-forward input $I_{\text{ff},i}^t$ and a recurrent input $I_{\text{rec},i}^t$ given by
\begin{equation}
\label{eq:current}
    I_i^t = I_{\text{ff},i}^t + I_{\text{rec},i}^t  + b_i = \mathbf{w}_{\text{ff},i}^\intercal \mathbf{x}^t  + \mathbf{w}_{\text{rec},i}^\intercal \mathbf{y}^{t-1}  + b_i,
\end{equation}
with feed-forward weight vector $\mathbf{w}_{\text{ff},i} \in \mathbb{R}^d$, recurrent weight vector $\mathbf{w}_{\text{rec},i}\in \mathbb{R}^m$, scalar bias $b_i$ and input vector $\mathbf{x}^t \in \mathbb{R}^d$, given by the $t$-th element of some (possibly infinitely long) input time series $X = [\mathbf{x}^1, \mathbf{x}^2, \ldots]$. Vector $\mathbf{y}^{t-1} \in \mathbb{R}^m$ contains the outputs $y_j^{t-1}$ of all neurons $j \in \{1,\ldots,m\}$ of the same layer from the previous time step.
\subsection{Approximate forward propagation}
In e-prop \citep{bellecSolutionLearningDilemma2020}, parameter updates are calculated online in the sense that no back-propagation of gradients is required. Let $\theta_i \in \mathbb{R}^p$ denote the vector of all parameters of neuron $i$ and let $\mathcal{L}^t$ denote some loss at time step $t$. In each time step $t$, a corresponding approximate parameter gradient (APG) $\apg \approx \nabla_{\theta_i} \mathcal{L}^t$ is computed and passed to an optimizer such as ADAM \citep{kingma2017adam} to obtain parameter updates. The APG $\apg$ is given by
\begin{equation}
\label{eq:eprop_weight_update}
    \apg =  \frac{d \mathcal{L}^t}{d y_i^t}\frac{\partial y_i^t}{\partial \mathbf{s}^t_i}\left[\frac{d\mathbf{s}_i^t}{d\theta_i}\right]_\text{local},
\end{equation}
where $\left[\fraci{d\mathbf{s}_i^t}{d\theta_i}\right]_\text{local}$ is a local approximation of the gradient of the neuron state $\mathbf{s}_i^t$ with respect to its parameters, see Appendix \ref{app:bptt} for its exact definition. Here and in the following, we use the notation $\partial$ for partial derivatives, and the notation $d$ for total derivatives, in line with \cite{bellecSolutionLearningDilemma2020}. The full gradient $\fraci{d\mathbf{s}_i^t}{d\theta_i}$, as used in BPTT, contains all pathways by which $\theta_i$ directly as well as indirectly influences state $\mathbf{s}_i^t$. The indirect pathways emerge for example from the dependence of $\mathbf{s}_i^t$ on outputs $y_j^{t-1}$ from other neurons, which again depend on previous outputs 
%$y_i^{t-2}$ 
from neuron $i$ (see Fig.~\ref{fig:neuron_model}). In e-prop, these pathways are disregarded and $\left[\fraci{d\mathbf{s}_i^t}{d\theta_i}\right]_\text{local}$ contains only the neuron-internal gradient pathways through the state-to-state derivatives $\fraci{\partial s_i^{q}}{\partial s_i^{q-1}}$ without propagating through recurrent intra-layer neuron connections (Fig.~\ref{fig:neuron_model}b). 

In e-prop, the term $\left[\fraci{d \mathbf{s}_i^t}{d\mathbf{\theta}_i}\right]_\text{local}$ from Eq.~\eqref{eq:eprop_weight_update} is called eligibility matrix $\mathbf{e}_{i}^t \in \mathbb{R}^{k\times p}$ 
%(also called sensitivity matrix in RTRL \citep{williamsLearningAlgorithmContinually1989a}) 
and is computed in a forward manner together with the neuron states during the regular forward pass:
\begin{equation}
\label{eq:elig_matrix}
 \left[\frac{d \mathbf{s}_i^t}{d \theta_i}\right]_\text{local} \myeq \mathbf{e}_{i}^t = \frac{\partial \mathbf{s}^t_i}{\partial \mathbf{s}^{t-1}_i}\mathbf{e}_{i}^{t-1} + \frac{\partial \mathbf{s}^t_i}{\partial \theta_i}.
\end{equation}
From this eligibility matrix, the APG $\apg$ at time step $t$ can be computed online without a backward pass 
%new
by replacing $\left[\fraci{d \mathbf{s}_i^t}{d \theta_i}\right]_\text{local}$ with the forward-propagated eligibilities $\mathbf{e}_{i}^t$ in Eq.~\eqref{eq:eprop_weight_update}.

In an online learning algorithm such as e-prop, APGs are computed in a time-local manner; thus, the parameter update is computed directly at time $t$.
Because APGs are computed online in tandem with states and outputs, recurrent networks can be trained on arbitrarily long time series --- retaining potentially infinite training context via eligibility matrices \(\mathbf{e}_{i}^t\), depending on the timescale of the neuron dynamics. This is in contrast to truncated BPTT, which uses a strictly limited context window.
\subsection{Parallelization}
Training of recurrent models can be sped up drastically by parallelization of computations over time.
In SSMs, parallelization of both the forward and backward pass is made possible by choosing the state-to-state transition to be linear \citep{guCombiningRecurrentConvolutional2021}. In our notation, this would require a linear state transition function $f$ (Eq.~\eqref{eq:state_transition}) as well as a linear output function $g$ if recurrent connections are used. Hence, SSMs either use linear recurrent interactions \citep{voelkerLegendreMemoryUnits2019b} or no recurrent connections at all \citep{orvietoResurrectingRecurrentNeural2023,guMambaLinearTimeSequence2023a}. 

In this case, the network dynamics can be written as $\mathbf{s}^{t} = A^t \mathbf{s}^{t-1} + \mathbf{x}^t$, where $\mathbf{s}^{t}$ is the state vector (of the whole layer) at time $t$, $\mathbf{x}^t$ is the input vector, and $A^t$ denotes the time-variant state transition matrix.
Then, the series of states $[\mathbf{s}^1, \mathbf{s}^2, \mathbf{s}^3, \ldots]$ can be written explicitly as $[\mathbf{x}^1, A^1 \mathbf{x}^1 + \mathbf{x}^2, A^2A^1 \mathbf{x}^1 + A^2 \mathbf{x}^2 + \mathbf{x}^3, \ldots]$ which can be efficiently computed via the associative scan (also called parallel prefix sum) algorithm \citep{BlellochTR90} in $\mathcal{O}(\log T)$ time where $T$ is the sequence length. Please refer to \cite{smithSimplifiedStateSpace2023a} for a more detailed explanation.

Obviously, this parallelization is not possible for usual RSNNs, since the spiking mechanism is by definition nonlinear. Nevertheless, we show below that training with approximate forward propagation can be parallelized even in nonlinear RSNNs.
\section{Hybrid propagation (HYPR)}
\label{sec:hypr}
\subsection{Parallelization in forward gradient learning}
To illustrate the relationship between forward gradient learning and parallelizable SSMs, we reformulate the computation of eligibility matrices $\mathbf{e}_{i}^t \in \mathbb{R}^{k \times p}$ from Eq.~\eqref{eq:elig_matrix} and APGs from Eq.~\eqref{eq:eprop_weight_update} into a linear SSM form:
\begin{align}
\label{eq:elig_vector_ssm}
\mathbf{e}_{i}^t &=  A_i^t\mathbf{e}_{i}^{t-1} + \delta_{i}^t \\
\label{eq:elig_vector_ssm2}
\apg &= B_i^t\mathbf{e}_{i}^t, 
\end{align}
with $A_i^t = \frac{\partial \mathbf{s}^t_i}{\partial \mathbf{s}^{t-1}_i}$, $\delta_{i}^t = \frac{\partial \mathbf{s}^t_i}{\partial \theta_{i}}$ and $B_i^t=\frac{d \mathcal{L}^t}{d y_i^t}\frac{\partial y_i^t}{\partial \mathbf{s}_i^t}$.
Remarkably, this formulation is linear despite possible non-linearities in the functions $f$ and $g$ (Eq.~\eqref{eq:state_transition}) of the neuron model at hand. These non-linearities are implicitly contained in the partial derivatives 
$\fraci{\partial \mathbf{s}^t_i}{\partial \mathbf{s}^{t-1}_i}$ 
and 
$\fraci{\partial y_i^t}{\partial \mathbf{s}_i^t}$, 
which are --- by definition of gradients --- linear first-order approximations. The linear SSM from Eqs.~\eqref{eq:elig_vector_ssm} and \eqref{eq:elig_vector_ssm2} can be either solved recurrently step-by-step (as in e-prop), or more efficiently for multiple time steps in parallel, as the SSM literature suggests \citep{smithSimplifiedStateSpace2023a}. The latter is the heart of our HYPR algorithm: We can overcome the sequentiality bottleneck of e-prop by exploiting the associativity of these linear operations. This applies to both: We can calculate --- in parallel --- the updates to eligibility matrices over multiple time steps and combine them at the end, as well as time step-wise APGs over multiple time steps which we combine at the end. The thereby obtained cumulative APGs are equivalent to the cumulative APGs in e-prop, but computed orders of magnitudes faster. In Section \ref{sec:experiments} we report a speedup of $108 \times$ for a medium size SNN with $1$M parameters. 

\begin{wrapfigure}{R}{0.61\textwidth}
\vspace{-12pt}
\begin{minipage}{0.60\textwidth}
\begin{algorithm}[H]
\caption{HYPR}
\label{alg:overview}
\SetAlgoLined
\KwIn{Time series $X$}
\KwIn{Network $\mathcal{N}$ with parameters $\theta$}
 $\mathbf{e}_{\theta}^{0} \leftarrow \mathbf{0}$\;
 $\cumapg \leftarrow \mathbf{0}$\;
\ForEach{Subsequence $\bar{X}_l$ in time series $X$}{
$\mathbf{s}^{1...\lambda}$, $\mathbf{y}^{1...\lambda}$, $\mathcal{L}^{1...\lambda}$ $\leftarrow$ \texttt{S-stage($\mathcal{N}$, $\bar{X}_l$)}\; 
    $\mathbf{e}_{\theta}^{\lambda}$, $[\apgnotnoi]^{1:\lambda}$ $\leftarrow$ \texttt{P-stage($\mathcal{N}$, $\bar{X}_l$, $\mathbf{s}^{1...\lambda}$,  \\
    \hspace{11.5em}$\mathbf{y}^{1...\lambda}$, $\mathbf{e}_{\theta}^{0}$, $\mathcal{L}^{1...\lambda}$)}\;   
    $\mathbf{e}_{\theta}^{0} \leftarrow \mathbf{e}_{\theta}^{\lambda}$\;
    $\cumapg\leftarrow \cumapg+[\apgnotnoi]^{1:\lambda}$\;
    }
$\theta\leftarrow \texttt{optimizer}\left(\cumapg\right)$\;
\end{algorithm}
\end{minipage}
\vspace{-10pt}
\end{wrapfigure}

In HYPR, time series are processed in segments which we refer to as subsequences. HYPR traverses through theses subsequences in their temporal order, computes parameter updates within the subsequence efficiently, and propagates eligibility matrices to the next subsequence.
For each subsequence, HYPR can be separated into two stages: a \textit{sequential S-stage}, where the subsequence is passed through the network sequentially item-by-item, and a \textit{parallel P-stage}, in which the APGs as well as eligibility matrices for further forward propagation to successive subsequences are computed in parallel over the time dimension of the subsequence.

More formally, let $X = \mathbf{x}^{1\ldots T}$ denote the input sequence. Here and in the following, we use the notation $\mathbf{x}^{1\ldots T}$ to denote a sequence $[\mathbf{x}^{1}, \ldots , \mathbf{x}^T]$.
First, $X$ is split into subsequences $\bar{X}_l=\bar{\mathbf{x}}_{l}^{1\ldots \lambda}$ of length $\lambda$ for $l\in\{1,2,\dots\}$. Hence, the $t$-th item $\bar{\mathbf{x}}_{l}^t$ in subsequence $\bar{X}_l$ corresponds to item $\mathbf{x}^{\lambda (l-1)+t}$ in $X$. 
For ease of notation, we explain HYPR with respect to one specific subsequence $\bar{X}_l$ and drop the subsequence index $l$. A brief overview of the algorithm is shown in Algorithm \ref{alg:overview}, for detailed pseudo-code refer to Algorithm \ref{alg:pseudocode} in Appendix \ref{app:pseudocode}.

During the S-stage of HYPR, we sequentially compute neuron states $\mathbf{s}_i^{1...\lambda}$, outputs  $y_i^{1...\lambda}$, as well as losses $\mathcal{L}^{1...\lambda}$ of the network $\mathcal{N}$ over subsequence $\bar{X}$:
\begin{align}
\label{eq:s_stage}
    \mathbf{s}_i^{1...\lambda}, y_i^{1...\lambda}, \mathcal{L}^{1...\lambda}  &= \texttt{S-stage}(\mathcal{N}, \bar{X}).
\end{align}  
The sequentiality of the S-stage cannot be avoided since we assume an arbitrary non-parallelizable neuron model, for example a vanilla LIF neuron.
However, we significantly reduce the computational burden of this sequential forward pass by postponing the computation of eligibility matrices and APGs to the later, parallel \texttt{P-stage}.
The neuron states and outputs obtained during the \texttt{S-stage} are then cached for the \texttt{P-stage}. Note, that the required memory is independent of the length of the original (potentially infinitely long) sequence, but is rather $\mathcal{O}\left(\lambda\right)$. I.e., it scales linearly with hyperparameter $\lambda$, the subsequence length, which can be chosen with respect to the available memory.

In the P-stage of the HYPR algorithm, eligibility matrices and APGs are computed efficiently in parallel. We can summarize the P-stage of HYPR as calculating the eligibility matrix $\mathbf{e}_{i}^{\lambda}$ at the end of the subsequence, as well as the cumulative APG $[\apgnot]^{1:\lambda}= \sum_{t=1}^\lambda \apg$ over the subsequence:
\begin{align}
\label{eq:hypr_routine}
    \mathbf{e}_{i}^{\lambda}\;, [\apgnot]^{1:\lambda}  &= \texttt{P-stage}(\mathbf{e}_{i}^{0}\;,  \bar{X}, I_i^{1\ldots\lambda}\;, y_i^{1\ldots\lambda}, \mathbf{s}_i^{1\ldots\lambda}),
\end{align} 
where $\mathbf{e}_{i}^{0}$ denotes the eligibility matrix from the end of the previous subsequence.
In Section \ref{subsec:elig} we describe how HYPR computes $\mathbf{e}_{i}^{\lambda}$, and in Section  \ref{subsec:weight_updates} we describe how it obtains $[\apgnot]^{1:\lambda}$. 
\subsection{Efficient calculation of $\mathbf{e}_{i}^{\lambda}$}
\label{subsec:elig}
We can unroll the recurrent definition of $\mathbf{e}_{i}^{t}$ from Eq.~\eqref{eq:elig_vector_ssm} to obtain the explicit representation
\begin{align}
\label{eq:elig_unrolled}
    \mathbf{e}_{i}^{\lambda} = \delta_{i}^\lambda + \underbrace{A_i^\lambda}_{\phi_i^{\lambda:\lambda}} \delta_{i}^{\lambda-1} + \underbrace{A_i^\lambda A_i^{\lambda-1}}_{\phi_i^{\lambda:\lambda-1}} \delta_{i}^{\lambda-2}  + \ldots + \underbrace{A_i^\lambda \ldots  A_i^{2}}_{\phi_i^{\lambda:2}} \delta_{i}^{1}  + \underbrace{A_i^\lambda \ldots  A_i^{1}}_{\phi_i^{\lambda:1}} \mathbf{e}_{i}^{0}.
\end{align}
Partial derivatives for neuron parameters 
$\delta_{i}^t = \fraci{\partial \mathbf{s}^t_i}{\partial \theta_{i}}$
are trivial to obtain in parallel as shown in Appendix \ref{app:param_grads}. State-to-state derivatives $A^t$ can also be obtained in parallel as shown in Appendix \ref{app:partial_grads}.
We define each $\phi_i^{\lambda:t}$ as the cumulative state transition matrix from $t$ to $\lambda$, given by
\begin{align}
    \phi_i^{\lambda:t} = \prod_{k=\lambda}^t {A_i^k} = \prod_{k=\lambda}^t \frac{\partial \mathbf{s}^k_i}{\partial \mathbf{s}^{k-1}_i}.
\end{align}
Matrices $\phi_i^{\lambda:t}$ can be computed in parallel with time complexity $\mathcal{O}(\log \lambda)$ and memory complexity $\mathcal{O}(\lambda)$ for all time steps using the associative scan algorithm. 
Finally, we can calculate $\mathbf{e}_{i}^{\lambda}$ as
\begin{equation}
    \label{eq:parallelizable_elig}
    \mathbf{e}_{i}^{\lambda} =  \delta_{i}^\lambda+ \sum_{t=1}^{\lambda-1} \delta_{i}^t \phi^{\lambda:t+1} + \mathbf{e}_{i}^{0}  \; \phi^{\lambda:1}.
\end{equation}
Note that Eq.~\eqref{eq:parallelizable_elig} alleviates the need to calculate the intermediate eligibility matrices $\mathbf{e}_{i}^{1}, \dots, \mathbf{e}_{i}^{\lambda-1}$.
The direct computation of the final $\mathbf{e}_{i}^{\lambda}$ from the sequence $\left[\delta_{i}^1, \delta_{i}^2, \ldots, \delta_{i}^\lambda\right]$ and $\mathbf{e}_{i}^{0}$ provides one of the major sources of efficiency gain in HYPR. The reasons are two-fold: First, the operation from Eq.~\eqref{eq:parallelizable_elig} can be fully parallelized on a GPU, since it is explicit and all individual terms are independent of each other. Second, in vanilla e-prop, the entire eligibility matrix (which is relatively large) is loaded from memory, updated, and stored in memory at each time step as shown in Eq.~\eqref{eq:elig_matrix}, resulting in significant memory I/O. In HYPR we exploit that the terms $\delta_{i}^t$ are outer products of two vectors (see Appendix \ref{app:param_grads}).
Hence, it is much more efficient to first collect the low-rank factors of these terms for all time steps and only then update the eligibility matrix by the efficient operation from Eq.~\eqref{eq:parallelizable_elig}. Intuitively, this can be interpreted as simultaneously forward-projecting all low-rank intermediate eligibility matrix updates and then combining them in the final time step via the sum-term in Eq.~\eqref{eq:parallelizable_elig}, instead of step-by-step propagating the full eligibility matrix. 
\subsection{Putting it all together: Efficient calculation of $[\apgnot]^{1:\lambda}$}
\label{subsec:weight_updates}
Cumulative APGs $[\apgnot]^{1:\lambda}$ are computed efficiently by using a hybrid of a parallelized backward gradient accumulation through the subsequence and forward propagation of eligibility matrices from the previous subsequence.
This way, HYPR exploits the low-dimensional intermediate terms similar as in backpropagation, but still operates in a constant memory complexity regime and can hence be applied to infinitely long sequences. The constant memory complexity stems from the fixed length $\lambda$ of the subsequence of which we compute the APGs. 

Let $\mathcal{L} = \sum_{t=1}^{\lambda} \mathcal{L}^t$ denote a summative loss function, where component $\mathcal{L}^t$ can be obtained from the network output $\mathbf{y}^t$ at time $t$. 
Consider the cumulative APG $[\apgnot]^{1:\lambda} = \sum_{t=1}^\lambda \apgnot^t$, which can be explicitly written as (see Eq.~\eqref{eq:elig_vector_ssm2})
\begin{equation}
    [\apgnot]^{1:\lambda} = \frac{d \mathcal{L}^\lambda}{d \mathbf{s}_i^\lambda}\mathbf{e}_{i}^{\lambda} + \frac{d \mathcal{L}^{\lambda-1}}{d \mathbf{s}_i^{\lambda-1}}\mathbf{e}_{i}^{\lambda-1}  + \ldots + \frac{d \mathcal{L}^1}{d \mathbf{s}_i^1}\mathbf{e}_{i}^{1}.
\end{equation}
We can unroll each eligibility matrix $\mathbf{e}_{i}^{t}$ according to Eq.~\eqref{eq:elig_unrolled} and reorder the terms as shown in Eq.~\eqref{eq:cum_weight_upd_decomposed} in Appendix \ref{app:backward_ssm} to obtain
\begin{equation}
    \label{eq:updates_hypr}
    [\apgnot]^{1:\lambda} = \mathbf{q}_i^\lambda \delta_{i}^\lambda\ldots + \mathbf{q}_i^3 \delta_{i}^3 + \mathbf{q}_i^2 \delta_{i}^2 + \mathbf{q}_i^1 \delta_{i}^1 + \mathbf{q}_i^0 \mathbf{e}_{i}^{0} =  \mathbf{q}^0_i \; \mathbf{e}_{i}^{0}  + \sum_{t=1}^\lambda \mathbf{q}_i^t \delta_{i}^t,
\end{equation}
where the vectors $\mathbf{q}_i^t$ are given by
\begin{equation}
    \mathbf{q}_i^t =  \frac{d \mathcal{L}^{\lambda}}{d \mathbf{s}_i^{\lambda}}A_i^{\lambda}A_i^{\lambda-1}\ldots A_i^{t+1} + \ldots + \frac{d \mathcal{L}^{t+2}}{d \mathbf{s}_i^{t+2}}A_i^{t+2}A_i^{t+1} +\frac{d \mathcal{L}^{t+1}}{d \mathbf{s}_i^{t+1}}A_i^{t+1} + \frac{d \mathcal{L}^t}{d \mathbf{s}_i^t}. 
    \label{eq:backward_ssm}
\end{equation}
This equation can again be written recursively in linear SSM form
\begin{equation}
\label{eq:q_recursive}
    \mathbf{q}_i^t =  \mathbf{q}_i^{t+1}A_i^{t+1}  +  \frac{d \mathcal{L}^t}{d \mathbf{s}_i^t},
\end{equation}
which can, similar to how we solve the linear SSM form from Eq.~\eqref{eq:elig_vector_ssm}, be parallelized (see Appendix \ref{app:backward_hypr} for details).
The key observation here is that we can accelerate the calculation of coefficients $q_i^t$ for all time steps $t \in \{ 1,\ldots,\lambda \}$ using the parallel associative scan algorithm. 
HYPR computes cumulative APGs via Eq.~\eqref{eq:updates_hypr} by combining vectors $\mathbf{q}_i^{0\ldots\lambda}$, gradients $\delta_{i}^{1\ldots \lambda}$, and the previous eligibility matrix $\mathbf{e}_{i}^{0}$ (see Appendix~\ref{app:schema_illustration} for an illustration). The resulting APGs and parameter updates are equivalent to those from the fully-forward variant ($\lambda=1$), but are computed more efficiently (see Sec.~\ref{sec:exp_memory_time}), making training independent of $\lambda$. 
Looking at Eq.~\eqref{eq:backward_ssm}, it may seem that APGs $\apg$ depend on future losses as in BPTT, which would contradict online learning.
This is not the case since HYPR is mathematically equivalent to fully-online e-prop (see Appendix~\ref{app:backward_ssm}).
Our narrative was so far based on a single recurrent layer network. 
See Appendix \ref{app:multi_layer} for the multi-layer case.
\section{Experiments}
\label{sec:experiments}
\subsection{Comparing time- and memory-requirements of HYPR and BPTT}
\label{sec:exp_memory_time}
As an initial experiment, we compared HYPR and BPTT in terms of memory and runtime performance on long sequences. 
As a network, we used an RSNN with a single hidden layer consisting of 1024 Balanced Resonate and Fire (BRF)~\citep{higuchi2024balanced} neurons, followed by an output layer of leaky integrator (LI) neurons ($\approx 1$M parameters). We tested the network on a toy task, to which we refer as the cue task (see Appendix \ref{app:details_datasets} for details).
\begin{figure}[!t]
    \centering
    \includegraphics[width=\linewidth]{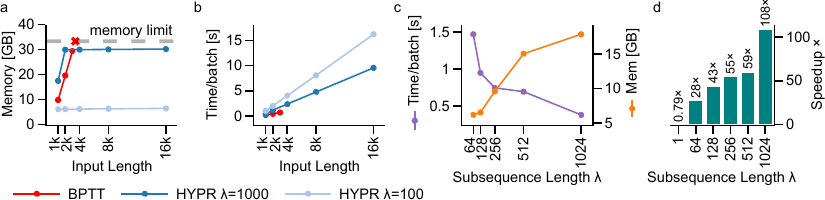}    
    \caption{Comparison of execution time and memory usage for BPTT and HYPR on the cue task.
    \textbf{a} GPU memory versus input length for BPTT and HYPR. The symbol $\mathbf{\times}$ denotes the estimated input length ($3{,}400$) at which BPTT requires more memory than available. If the subsequence length $\lambda$ is equal to the total input sequence length, reduced memory consumption of HYPR can be noticed (HYPR with $\lambda=1000$ at $1$k input length), due to internal optimizations of the compiler. 
    \textbf{b} Training time for a single batch.
    \textbf{c} Memory and training time of HYPR with different subsequence lengths $\lambda$ for an input of length $1024$. 
    \textbf{d} Speedup of HYPR compared to e-prop.
    }
    \label{fig:profiling}
\end{figure}
The network received inputs from 15 spiking input neurons. In the first 20 time steps, either neurons 1 to 5 (input class A) or neurons 6 to 10 (input class B) were active, while other neurons remained silent. This cue was followed by a (potentially very long) delay period where all input neurons remained silent. At the end of the sequence, input neurons 11 to 15 became active, indicating a recall period. The network output should then indicate during the recall period whether the initial cue belonged to class A or B. This task tests the long-term credit assignment capabilities of a learning algorithm: Only if the information provided at the first cue is successfully propagated to the end of the sequence, the task can be learned. An advantage of this task is the possibility to directly control the input sequence length and with it the length of long-term dependencies in the task. 
In Fig.~\ref{fig:profiling} we compare the memory consumption and wall clock training time for HYPR and BPTT on this tasks. For all input lengths shown in Fig.~\ref{fig:profiling} the trained networks were able to solve the task (training accuracy 100\%), confirming successful temporal credit assignment with both algorithms, BPTT and HYPR. 
For sequence lengths of above approximately $3{,}400$, BPTT required more memory than available on the tested GPU, whereas HYPR consumes constant memory, regardless of the sequence length.  With parallelization we could achieve up to a $108\times$ speedup over the sequential e-prop, see Fig.~\ref{fig:profiling}d. 
Since the memory consumption of HYPR does not scale with input sequence length (given constant $\lambda$), it can work with arbitrary-length sequences on a single GPU without exceeding its memory. 
\subsection{Narrowing the performance gap between approximate forward propagation and BPTT}
\label{sec:exp_performance}
\begin{table}[!t]
\caption{Comparison of BPTT and HYPR learning algorithms on benchmark datasets. Here and in the following tables, results are reported on the test set, and statistics (mean $\pm$ std.~dev.) were performed over 5 random seeds, where the parameter initialization and sampling of the validation set for model selection are randomized each time. BPTT entries include the original accuracies reported by the respective authors in parentheses if available. The number of parameters depends on the model and dataset but was equivalent between HYPR and BPTT, see Appendix \ref{app:training_and_hyperparams}.} 
\vspace{.2cm}
\label{tab:benchmarks}
\scalebox{0.90}{
\begin{tabular}{l c lll}
\toprule
& & \textsc{SHD} & \textsc{ECG} & \textsc{sMNIST} \\
\midrule
\multirow{2}{*}{\parbox[t]{3.5cm}{\centering \textbf{ALIF}~\citep{yin2021accurateefficienttimedomainclassification}\\(Yin et al., 2021)}}
& {\parbox[t]{1.2cm}{\centering\textbf{BPTT}}} & 85.61 $\pm$ 1.67 \scalebox{0.9}{(90.4\footnotemark[1])} & 84.24 $\pm$ 0.75 \scalebox{0.9}{(85.9\footnotemark[1])} & 97.57 $\pm$ 0.21 \scalebox{0.9}{(98.7\footnotemark[1])} \\
& {\parbox[t]{1.2cm}{\centering\textbf{HYPR}}} & 85.33 $\pm$ 1.39 & 79.91 $\pm$ 0.74 & 82.50 $\pm$ 1.91 \\
\midrule
\multirow{2}{*}{\parbox[t]{3.5cm}{\centering \textbf{BRF}~\citep{higuchi2024balanced}\\(Higuchi et al., 2024)}}
& {\parbox[t]{1.2cm}{\centering\textbf{BPTT}}} & 92.61 $\pm$ 0.26 \scalebox{0.9}{(91.7)} & 85.77 $\pm$ 0.55 \scalebox{0.9}{(85.8)} & 98.91 $\pm$ 0.08 \scalebox{0.9}{(99.0)} \\
& {\parbox[t]{1.2cm}{\centering\textbf{HYPR}}} & 91.20 $\pm$ 0.51 & 83.70 $\pm$ 0.78 & 96.12 $\pm$ 2.34 \\
\midrule
\multirow{2}{*}{\parbox[t]{3.5cm}{\centering \textbf{SE-adLIF}~\citep{baronig2025advancingspatiotemporalprocessingspiking}\\(Baronig et al., 2025)}}
& {\parbox[t]{1.2cm}{\centering\textbf{BPTT}}} & 93.45 $\pm$ 0.96 \scalebox{0.9}{(93.8)} & 86.90 $\pm$ 0.53 \scalebox{0.9}{(86.9)} & 99.06 $\pm$ 0.05  \\
& {\parbox[t]{1.2cm}{\centering\textbf{HYPR}}} & 92.17 $\pm$ 0.39 & 83.38 $\pm$ 0.49 & 94.11 $\pm$ 0.52 \\
\bottomrule
\end{tabular}}
\end{table}
\footnotetext[1]{We were not able to exactly reproduce these reported results for multiple seeds with our code. However, since our goal was to compare the relative performance of BPTT and HYPR, this discrepancy is not critical.}

In HYPR, APGs are computed based on approximate gradients of the parameters. This approximation might affect task performance compared to BPTT. To investigate on this matter, we compared both algorithms on benchmark datasets  commonly used in the SNN and RNN research communities: Spiking Heidelberg Digits (SHD) \citep{cramerHeidelbergSpikingData2022}, an ECG dataset~\citep{laguna1997database}, and sequential MNIST (sMNIST) \citep{bellec2018long,lecun1998mnist}. We compared the accuracy of BPTT and HYPR on networks based on three different neuron models: two oscillatory models, BRF \citep{higuchi2024balanced} and SE-adLIF \citep{baronig2025advancingspatiotemporalprocessingspiking}, as well as ALIF \citep{yin2021accurateefficienttimedomainclassification}, a non-oscillatory neuron model with threshold adaptation, see Appendix \ref{app:neuron_models} for detailed descriptions of the neuron models. The latter model was initially used for testing the capabilities of the approximate forward propagation algorithm e-prop \citep{bellecSolutionLearningDilemma2020}. The results are shown in Tab.~\ref{tab:benchmarks}.
Due to the approximative nature of HYPR, we do not expect it to outperform BPTT on any benchmark. 
However, we can observe that across the wide variety of tasks, HYPR is mostly on par with BPTT for the two oscillatory neuron models, BRF and SE-adLIF. This result is surprising since in HYPR the gradient pathway through the recurrent synaptic connections is disregarded. We account the good performance of HYPR to the choice of neuron model: The spiking neuron models that currently achieve the best results on the benchmark datasets at hand are all of oscillatory nature, a feature that has recently been studied extensively %in both the SNN 
\citep{bal2025pspikessm,baronig2025advancingspatiotemporalprocessingspiking,bittar2022surrogate,higuchi2024balanced}. 
These oscillations support propagation of time-sensitive information through their state-to-state transitions, a feature that might naturally be well exploited by the forward gradient learning of HYPR. We also found that networks trained with HYPR benefit from additional layers, see Appendix \ref{app:multi_layer_results}. Note, that we only considered RSNNs that do not violate any fundamental constraints of neuromorphic hardware: Models including features like layer normalization, floating-point-value-based communication between neurons (for example through skip connections), temporal convolutions, or attention were intentionally excluded.

Encouraged by these results, we explored the limitations of HYPR and tested it on challenging tasks with long-range dependencies from the long-range arena benchmarks~\citep{tay2020long}. 
We tested sequential CIFAR (sCIFAR) and the easier variant of Pathfinder, to which we refer to as Pathfinder-E.
\begin{wraptable}{r}{6.7cm}
\vspace{-8pt}
\caption{Classification accuracy (mean $\pm$ std.~dev.~over 5 runs) of HYPR-trained RSNNs with BRF neurons on two tasks of the long-range arena benchmark~\citep{tay2020long}.}
\label{tab:lra}
\scalebox{0.95}{
\begin{tabular}{c cc}
\toprule
& {\parbox[t]{2.3cm}{\centering\textsc{sCIFAR}}} & {\parbox[t]{2.3cm}{\centering\textsc{Pathfinder-e}}} \\
\midrule
\textbf{BPTT} & 63.86 \scalebox{0.9}{$\pm$0.74} & 85.10 \scalebox{0.9}{$\pm$0.27} \\
\textbf{HYPR} & 57.90 \scalebox{0.9}{$\pm$0.51} & 65.48 \scalebox{0.9}{$\pm$2.45}\\
\bottomrule
\end{tabular}}
\vspace{-8pt}
\end{wraptable}
In sCIFAR, the pixels of images from the CIFAR-10 datasets are presented sequentially to the network, and the network has to classify the image category (sequence length 1024). In Pathfinder-E, an image of line drawings is presented sequentially pixel-by-pixel (sequence length 1024), and the network has to decide whether a starting point is connected by a line with an end point. 
In Tab.~\ref{tab:lra}, we report the first successful application of RSNNs with an approximate forward-learning algorithm on sCIFAR and Pathfinder-E.
Nevertheless, we observe a larger performance gap between BPTT and HYPR in these challenging tasks, which remains to be explored further.

\subsection{Influence of Recurrent Connections}
\label{sec:recurrent_conn}
In HYPR, the gradient path through the recurrent synaptic connections is ignored. This raises the question, whether networks trained with HYPR can successfully learn to utilize these recurrent synaptic pathways.
To answer this question, we compared the performance of our networks to the performance of networks without recurrent connections, see Tab.~\ref{tab:rec_vs_norec}. 
We found 
\begin{wraptable}{r}{8.8cm}
\vspace{-7pt}
\caption{Comparison of performance on SHD, sMNIST and sCIFAR on networks of BRF neurons~\citep{higuchi2024balanced} trained with and without recurrent connections using BPTT and HYPR.}
\label{tab:rec_vs_norec}
\scalebox{0.95}{
\begin{tabular}{c c ccc}
\toprule
& \multirow{2}{*}[1pt]{\parbox[t]{1cm}{\centering \textbf{Rec.}\\\textbf{Conn.}}} & \multirow{2}{*}[0pt]{\parbox[t]{1.4cm}{\centering \textsc{SHD}}} & \multirow{2}{*}[0pt]{\parbox[t]{1.4cm}{\centering \textsc{sMNIST}}} & \multirow{2}{*}[0pt]{\parbox[t]{1.4cm}{\centering \textsc{sCIFAR}}} \\ \\
\midrule
\multirow{2}{*}{\textbf{BPTT}}
& \cmark & 92.61 \scalebox{0.9}{$\pm$0.26} & 98.91 \scalebox{0.9}{$\pm$0.08} & 63.86 \scalebox{0.9}{$\pm$0.74} \\
& \xmark & 90.86 \scalebox{0.9}{$\pm$0.57} & 95.69 \scalebox{0.9}{$\pm$0.13} & 53.41 \scalebox{0.9}{$\pm$0.50} \\
\midrule
\multirow{2}{*}{\textbf{HYPR}}
& \cmark & 91.20 \scalebox{0.9}{$\pm$0.51} & 96.12 \scalebox{0.9}{$\pm$2.34} & 57.90 \scalebox{0.9}{$\pm$0.51} \\
& \xmark & 89.49 \scalebox{0.9}{$\pm$0.66} & 93.01 \scalebox{0.9}{$\pm$0.19} & 49.46 \scalebox{0.9}{$\pm$0.46} \\
\bottomrule
\end{tabular}
}
\vspace{-10pt}
\end{wraptable}
that performance drops significantly when recurrent connections are dropped, suggesting that HYPR does indeed utilize recurrent connectivity despite its ignorance to their gradient pathways. Interestingly, the performance drop is higher for both BPTT and HYPR on the more demanding sCIFAR data set, indicating the importance of recurrent interactions for harder tasks.
\section{Discussion}
\label{sec:discussion}
This work introduces HYPR, a scalable and efficient segment-wise online training algorithm applicable to almost arbitrary RSNNs and RNNs. Through parallelization over segments, we achieve significant training speedup, up to 108$\times$ in our experiments, compared to the mathematically equivalent fully-online algorithm e-prop \citep{bellecSolutionLearningDilemma2020}. In contrast to e-prop, where weight updates are calculated in a time step-wise online manner, the segment-wise parallelization in HYPR effectively utilizes GPU resources while still operating in a constant memory regime, without sacrificing the infinite training context of e-prop.

We demonstrated that HYPR excels if applied to oscillatory spiking neuron models, a new generation of models that has been recently demonstrated to be powerful on various benchmarks \cite{baronig2025advancingspatiotemporalprocessingspiking,bittar2022surrogate,higuchi2024balanced}. We first showed how HYPR can overcome the memory limitations of BPTT at a comparable scaling of the runtime (Fig.~\ref{fig:profiling}a,b). 
Second, we demonstrated that the synergy between HYPR and oscillatory neuron models significantly narrows the gap between BPTT and approximate forward gradient learning as compared to previously proposed neuron models with threshold adaptation \citep{bellecSolutionLearningDilemma2020}.
Further, we demonstrated that, despite ignoring gradient pathways through recurrent connections, HYPR utilizes these connections for more accurate classification performance (Tab.~\ref{tab:rec_vs_norec}).

\paragraph{Limitations and future work} HYPR is based on an approximate gradient computation that neglects gradient paths through recurrent connections. As such, its accuracies should be below those achievable by BPTT. We found that, despite working surprisingly well on the challenging datasets sCIFAR and Pathfinder-E, a performance gap between BPTT and HYPR still persists (Tab.~\ref{tab:lra}).
Future work could investigate methods that combine forward-propagated eligibilities with truncated backpropagated gradients through recurrent connections.

Since HYPR is a relatively complex learning algorithm, it is potentially hard to implement on neuromorphic hardware, where pure forward propagation may be preferable. In any case, we believe that the achievable speedups on standard GPU-based architectures can boost research on efficient learning algorithms for RSNNs, which has been hindered by the sequentiality bottleneck of forward propagation algorithms so far.

\newpage
\appendix
\setcounter{figure}{0}
\renewcommand{\thefigure}{A\arabic{figure}}
\setcounter{table}{0}
\renewcommand{\thetable}{A\arabic{table}}
\setcounter{equation}{0}
\renewcommand{\theequation}{A\arabic{equation}}
\renewcommand{\thefootnote}{\arabic{footnote}}
\setcounter{algocf}{0} % Reset if needed
\renewcommand{\thealgocf}{A\arabic{algocf}}
\section*{Appendix}
\section{Computational overhead of forward-propagating gradients}
\label{app:overhead}
A major disadvantage of forward-propagating gradients (e.g. as in e-prop) is a significant overhead in multiplication operations, compared to backpropagation. As a guiding example, consider the following computational chain: $\mathbf{s}^{0}=f(\mathbf{w})$, $\mathbf{s}^{t+1} = g(\mathbf{s}^t)$, $\mathcal{L} = h(\mathbf{s}^\ell)$, with some parameter $\mathbf{w} \in \mathbb{R}^{p}$, intermediate values $\mathbf{s}^t \in \mathbb{R}^k$ and scalar $\mathcal{L} \in \mathbb{R}$. Computing the gradient $\nabla_{\mathbf{w}} \mathcal{L}$ via the chain rule results in the following chain of multiplications:
\begin{equation}
    \nabla_{\mathbf{w}} \mathcal{L} =  
    \underbrace{\nabla_{\mathbf{s}^\ell} \mathcal{L}}_{\mathbb{R}^{k}} 
    \,
    \underbrace{\frac{\partial \mathbf{s}^{\ell}}{\partial  \mathbf{s}^{\ell-1}}}_{\mathbb{R}^{k \times k}}
    \,
    \underbrace{\frac{\partial \mathbf{s}^{\ell-1}}{\partial  \mathbf{s}^{\ell-2}}}_{\mathbb{R}^{k \times k}}
    \,
    \ldots
    \,
    \underbrace{\frac{\partial \mathbf{s}^{2}}{\partial  \mathbf{s}^{1}}}_{\mathbb{R}^{k \times k}}
    \,
    \underbrace{\frac{\partial \mathbf{s}^{1}}{\partial  \mathbf{s}^{0}}}_{\mathbb{R}^{k \times k}}
    \,
    \underbrace{\frac{\partial \mathbf{s}^{0}}{\partial \mathbf{w}}}_{\mathbb{R}^{k \times p}}  .
\end{equation}
One could resolve this chain in a forward manner, i.e. 
\begin{equation}
\nabla_{\mathbf{w}} \mathcal{L}
=
\nabla_{\mathbf{s}^\ell} \mathcal{L}
\,
\frac{\partial \mathbf{s}^\ell}{\partial \mathbf{s}^{\ell-1}} \Bigl(
\cdots
\Bigl(\frac{\partial \mathbf{s}^3}{\partial \mathbf{s}^2}
\,
    \underbrace{\Bigl(\frac{\partial \mathbf{s}^2}{\partial \mathbf{s}^1}
    \,
      \underbrace{\Bigl(\frac{\partial \mathbf{s}^1}{\partial \mathbf{s}^0}
\,
  \frac{\partial \mathbf{s}^0}{\partial \mathbf{w}}  
\Bigr)}_{\mathbb{R}^{k \times p}}\Bigr)\Bigr)
}_{\mathbb{R}^{k \times p}} \cdots \Bigr),
\end{equation}
where terms in brackets are evaluated first, or in a backward manner
\begin{equation}
\nabla_{\mathbf{w}} \mathcal{L}
=
\Bigl( \cdots \underbrace{  
  \underbrace{\Bigl(\Bigl(\nabla_{\mathbf{s}^\ell} \mathcal{L}
  \,
  \frac{\partial \mathbf{s}^\ell}{\partial \mathbf{s}^{\ell-1}}\Bigr)}_{\mathbb{R}^{k}}
  \,
    \frac{\partial \mathbf{s}^{\ell-1}}{\partial \mathbf{s}^{\ell-2}}\Bigr)}_{\mathbb{R}^{k}}
  \cdots
    \frac{\partial \mathbf{s}^1}{\partial \mathbf{s}^0} \Bigr)
    \,
    \frac{\partial \mathbf{s}^0}{\partial \mathbf{w}}.
\end{equation}
Although the result is equivalent, the intermediate terms in the forward manner are significantly larger and require a significantly higher amount of multiplication operations and memory compared to the backward manner. While the forward mode requires $\ell\,k^2p + k\,p$ multiplications, the backward mode only requires $\ell\,k^2 + k\,p$ multiplications. Usually, $p$ is the largest of the three variables ($k,\ell,p$) and in the order of millions. Hence, training algorithms involving forward propagation of gradients, for example RTRL \cite{williamsLearningAlgorithmContinually1989a} or e-prop \cite{bellecSolutionLearningDilemma2020}, require significantly higher memory I/O compared to BPTT since they forward-propagate and materialize large sensitivity/eligibility matrices in each time step.
\section{Direct and indirect gradient pathways}
\label{app:bptt}
In BPTT, the exact total derivatives $\frac{d \mathbf{s}_j^t}{d \theta_i}$ are used to compute the parameter update. Recursively written, they can be computed by
\begin{align}
\label{eq:exact_grad}
    \frac{d \mathbf{s}_j^t}{d \theta_i} &= \underbrace{\frac{\partial \mathbf{s}_j^t}{\partial \theta_i}  +  \frac{\partial \mathbf{s}_j^{t}}{\partial \mathbf{s}_j^{t-1}}\frac{d \mathbf{s}_j^{t-1}}{d \theta_i}}_{\substack{\text{direct} \\  \text{(neuron-internal recurrence)}}} + \underbrace{\sum_{k\neq j} \frac{\partial \mathbf{s}_j^t}{\partial \mathbf{y}_k^{t-1}} \frac{\partial \mathbf{y}_k^{t-1} }{\partial \mathbf{s}_k^{t-1}} \frac{d \mathbf{s}_k^{t-1}}{d \theta_i}}_{\substack{\text{indirect} \\  \text{(intra-layer recurrence)}}}
\end{align}
which consists of a direct gradient pathway (neuron-internal recurrence) that does not "leave" the neuron, and an indirect pathway (intra-layer recurrence) through other neurons, see also Fig.~\ref{fig:neuron_model}. In e-prop, $\fraci{d \mathbf{s}_j^t}{d \theta_i}$ is replaced by a local approximation $[\fraci{d \mathbf{s}_j^t}{d \theta_i}]_\text{local}$ with \mbox{$[\fraci{d \mathbf{s}_j^t}{d \theta_i}]_\text{local} \myeq \mathbf{0} \; \forall \; j\neq i$} which cancels all indirect components from Eq.~\eqref{eq:exact_grad}. The non-zero terms $[\fraci{d \mathbf{s}_i^t}{d \theta_i}]_\text{local}$ are defined recursively by
\begin{align}
    \left[\frac{d \mathbf{s}_i^t}{d \theta_i}\right]_\text{local} &= \frac{\partial \mathbf{s}_i^t}{\partial \theta_i}   + \frac{\partial \mathbf{s}_i^{t}}{\partial \mathbf{s}_i^{t-1}}\left[\frac{d \mathbf{s}_i^{t-1}}{d \theta_i}\right]_\text{local}.
\end{align}
\section{Pseudocode of the HYPR algorithm}
\label{app:pseudocode}
In Algorithm \ref{alg:pseudocode} we show pseudocode for the HYPR algorithm. The algorithm is applied to batches of data, we omit the batch dimension in the pseudocode for simplicity. We denote the parallel computation of Jacobians as $\texttt{jac}$, where the first argument corresponds to the function to be differentiated, and \texttt{assoc\_scan} as the associative scan function, where the first argument defines the binary associative operator used by the scan algorithm.
\begin{algorithm}[!ht]
\DontPrintSemicolon
\caption{single-layer HYPR}
\label{alg:pseudocode}
\KwIn{$X^{1:T}\in\mathbb{R}^{T\times d}$} 
\KwIn{Network $\mathcal{N}$ with parameters $\theta = \{W_{\mathrm{ff}} \in \mathbb{R}^{d \times m},\; W_{\mathrm{rec}} \in \mathbb{R}^{m \times m},\; b \in \mathbb{R}^m\}$} 
\KwIn{Subsequence length $\lambda$}
\KwOut{Updated parameters $\theta$}
% Initialization
$\mathbf e^0 \gets \mathbf{0}$ \tcp*{$\in \mathbb R^{k\times d_{\theta}}$ with no.~of parameters $d_{\theta}$ in $\mathcal{N}$}
$\cumapg \gets \mathbf{0}$\;
\For{$\ell \leftarrow 1$ \KwTo $T / \lambda$}{
  $t_0 \gets (\ell-1)\,\lambda$\; 
  $\bar X \gets X^{(t_0+1)\ldots (t_0+\lambda)}$ \tcp*{get next subsequence $\bar X \in\mathbb R^{\lambda\times d}$}
  $I_{\mathrm{ff}} \gets \bar X\,W_{\mathrm{ff}} + b$ \tcp*{$I_{\mathrm{ff}}\in\mathbb R^{\lambda\times m}$ (parallel)}
  % \;
  {\vspace{10pt}}
  \tcc{S-stage (sequential)}
    {\vspace{4pt}}
  Initialize $\mathbf{s}^0 \leftarrow \mathbf{0}, \mathbf{y}^0 \leftarrow \mathbf{0}$ \;
  \For{$t \leftarrow 1$ \KwTo $\lambda$}{
    $I_{\mathrm{rec}}^t \gets W_{\mathrm{rec}}\,y^{t-1}$ \tcp*{$\in\mathbb R^m$}
    $I^t \gets I_{\mathrm{ff}}^t + I_{\mathrm{rec}}^t$ \;
    $\mathbf{s}^t \gets f(\mathbf{s}^{t-1}, I^t)$ \;
    $\mathbf{y}^t \gets g(\mathbf{s}^t)$  \;
    $\mathcal{L}^t \leftarrow \texttt{loss}(y^t)$ \tcp*{supervised or unsupervised loss function}
  }
    {\vspace{10pt}}
  \tcc{P-stage (parallel)}
    {\vspace{4pt}}
  $\bigl[\frac{\partial \mathbf{s}^1}{\partial \mathbf{s}^0},\,\dots,\,\frac{\partial \mathbf{s}^\lambda}{\partial \mathbf{s}^{\lambda-1}}\bigr],\bigl[\frac{\partial \mathbf{s}^1}{\partial I^1},\,\dots,\,\frac{\partial \mathbf{s}^\lambda}{\partial I^{\lambda}}\bigr] \leftarrow \texttt{jac}(f, \mathbf{s}^{0\ldots\lambda-1}, I^{1\ldots\lambda})$ 
  \;
  \tcp*{$\frac{\partial \mathbf{s}^{t+1}}{\partial \mathbf{s}^t}\in\mathbb{R}^{m\times k \times k}$, $\frac{\partial \mathbf{s}^{t}}{\partial I^t}\in\mathbb{R}^{m\times k}$, see Appendix~\ref{app:partial_grads}}
  $\bigl[\frac{d \mathcal{L}^1}{d \mathbf{s}^1},\,\dots,\,\frac{d \mathcal{L}^\lambda}{d \mathbf{s}^{\lambda}}\bigr]\gets \texttt{jac}(\{g,\texttt{loss}\}, \mathbf{s}^{1\ldots\lambda}, \mathbf{y}^{1\ldots\lambda})$ 
  \;
  \tcp*{$\frac{\partial \mathcal{L}^{t}}{\partial \mathbf{s}^t}\in\mathbb{R}^{m\times k}$, see Appendix~\ref{app:partial_grads}}
  $\bigl[\phi^{\lambda:1},\ldots,\;\phi^{\lambda:\lambda}\bigr] \gets \texttt{assoc\_scan}(\times, \bigl[\frac{\partial \mathbf{s}^1}{\partial \mathbf{s}^0},\,\dots,\,\frac{\partial \mathbf{s}^\lambda}{\partial \mathbf{s}^{\lambda-1}}\bigr])$ 
  \;
  \tcp*{$\phi^{\lambda:t} \in \mathbb{R}^{m \times k \times k}$, see text}
  $\mathbf{q}^{0\ldots \lambda} \gets \texttt{reverse\_scan}(\bigl[\frac{\partial \mathbf{s}^1}{\partial \mathbf{s}^0},\,\dots,\,\frac{\partial \mathbf{s}^\lambda}{\partial \mathbf{s}^{\lambda-1}}\bigr], \bigl[\frac{d \mathcal{L}^1}{d \mathbf{s}^1},\,\dots,\,\frac{d \mathcal{L}^\lambda}{d \mathbf{s}^{\lambda}}\bigr])$ 
  \;
  \tcp*{$\mathbf{q}^{t} \in \mathbb{R}^{m \times k}$, see Appendix \ref{app:backward_ssm}}
      $\mathbf e^\lambda \gets \mathbf e^0\,\phi^{\lambda:1} + \sum_{t=1}^{\lambda-1} \delta^t\,\phi^{\lambda:t+1} + \delta^\lambda$ 
      \;
      \tcp*{with $\delta^t=\frac{\partial \mathbf{s}^\lambda}{\partial I^\lambda}\frac{\partial I^t}{\partial \theta}$, decomposed as described in Appendix \ref{app:param_grads}}
  $[\tilde{\nabla} \theta]^{1:\lambda} \gets \mathbf{e}^0\delta^0 + \sum_{t=1}^\lambda \mathbf{q}^t \delta^t$ \;
    {\vspace{12pt}}
    $\mathbf e^0 \gets \mathbf e^\lambda$ \;
  $\cumapg \gets \cumapg + [\tilde{\nabla} \theta]^{1:\lambda}$ \;
}
$\theta \gets \texttt{optimizer}(\cumapg)$\;
\Return $\theta$ \;
\end{algorithm}
\section{Parameter gradients $\delta_{i}^t$}
\label{app:param_grads}
We intentionally separated the calculation of input $I_i^t$ given by Eq.~\eqref{eq:current} from the state transition $f$ in Eq.~\eqref{eq:state_transition} to emphasize that partial derivatives $\delta^t$ can be factorized into low-dimensional factors to significantly reduce the memory consumption and memory I/O of HYPR. By using these low-dimensional factors, the full parameter gradients $\delta^t$ are never materialized, enhancing performance and memory-efficiency of HYPR: For example for the feed-forward weight matrix $W_{\text{ff}} \in \mathbb{R}^{m \times d}$, parameter gradient $\delta^t_{W_{\text{ff}}}$ is given by the full Jacobian tensor $\frac{\partial S^t}{\partial W_{\text{ff}}} \in \mathbb{R}^{m \times k \times m \times d}$ with respect to the matrix $S^t \in \mathbb{R}^{m \times k}$ of all neuron states in time step $t$  and requires $\mathcal{O}(m^2 k d)$ memory. More specifically, its components are given as \mbox{$\bigl(\frac{\partial S^t}{\partial W_{\mathrm{ff}}}\bigr)_{ij\ell m}\;=\;\frac{\partial S^t_{ij}}{\partial W_{\mathrm{ff},\ell m}}$}. However, we found that this memory requirement can be drastically reduced by instead computing and caching some smaller matrices $D^t \in \mathbb{R}^{m \times k}$ with $D^t_{ij} = \fraci{\partial S^t_{ij}}{\partial I_{\text{ff},i}^t}$ together with vectors $\bar{\mathbf{x}}^t \in \mathbb{R}^d$ from subsequence $\bar{X}$. Note, that $\fraci{\partial I_{\text{ff},i}^t}{\partial W_{\text{ff},i}} = \bar{\mathbf{x}}^t$ for all neurons $i$, since all neurons receive the same input.
Together $D^t$ and $\bar{\mathbf{x}}^t$ consume only $\mathcal{O}(mk + d)$ memory and allow to efficiently perform the operations from Eq.~\eqref{eq:parallelizable_elig} and \eqref{eq:updates_hypr} using Einstein summations without ever materializing the large parameter gradient $\delta^t_{W_{\text{ff}}}$. 

Hence, we can express the components of $\delta^t$  that correspond to the parameters in the feed-forward weight matrix $W_{\text{ff}}$ in Eqs.~\eqref{eq:parallelizable_elig} and \eqref{eq:updates_hypr}  as
\begin{align}
    (\delta^t_{W_{\text{ff}}})_{ij\ell m} =  \biggl(\frac{\partial S^t}{\partial W_{\mathrm{ff}}}\biggr)_{ij\ell m} = \frac{\partial S^t_{ij}}{\partial W_{\mathrm{ff},\ell m}}  = \frac{\partial S_{ij}^t}{\partial I_{\text{ff},i}^t}\frac{\partial I_{\text{ff},i}^t}{\partial W_{\text{ff},im}^t} &=    \left\{
\begin{array}{ll}
D^t_{ij} \:\bar{\mathbf{x}}^t_m  & \text{if} \quad \ell=i \\
0 & \text{if} \quad \ell\neq i \\
\end{array}
\right. 
\end{align}
The same principle applies to $W_{\text{rec}}$, where $(\delta^t_{W_{\text{rec}}})_{ij\ell m} = D^t_{ij} \: \mathbf{y}^t_m   \;\; \text{if} \;\; \ell=i, \; \text{else} \; 0$. All matrices $D^t$ can be obtained in parallel as explained in Appendix \ref{app:partial_grads}.
\section{Efficient calculation of partial gradients}
\label{app:partial_grads}
All partial gradients involved in HYPR can be calculated efficiently in parallel. 
We first compute Jacobian matrices $J_{f,i}^t$ and $J_{g,i}^t$ of functions $f$ and $g$ from Eq.~\eqref{eq:state_transition}. The Jacobians are given by
\begin{equation}
\label{eq:jacobians}
    J_{f,i}^t = \begin{bmatrix}
         \frac{\partial s^t_{i,1}}{\partial s^{t-1}_{i,1}} & \dots & \frac{\partial s^t_{i,1}}{\partial s^{t-1}_{i,k}} & \frac{\partial s^t_{i,1}}{\partial I^{t}_{i}} \\
          \vdots & &  & \vdots  \\
         \frac{\partial s^t_{i,k}}{\partial s^{t-1}_{i,1}} & \dots & \frac{\partial s^t_{i,k}}{\partial s^{t-1}_{i,k}} & \frac{\partial s^t_{i,k}}{\partial I^{t}_{i}}
    \end{bmatrix} \in \mathbb{R}^{k \times (k+1)}, \qquad   J_{g,i}^t = \begin{bmatrix}
         \frac{\partial y^t_{i}}{\partial s^{t}_{i,1}}  \\
         \vdots \\
         \frac{\partial y^t_{i}}{\partial s^{t}_{i,k}}
    \end{bmatrix} \in \mathbb{R}^{k \times 1},
\end{equation}
where we recall that $k$ is the neuron state dimension. $s^t_{i,j}$ refers to the $j$-th state of neuron $i$ at time step $t$ and $I^{t}_{i}$ the input to neuron $i$ at time step $t$.
Given the cached latent states and outputs obtained in the S-stage, these Jacobians can be computed for all time steps of the subsequence in parallel. This operation has time complexity $\mathcal{O}(1)$, given sufficiently many concurrent processors. These Jacobian matrices can be obtained by using auto-differentiation frameworks, without requiring a model-dependent explicit implementation, referred to as \texttt{jac} in Algorithm \ref{alg:pseudocode}. This allows HYPR to be trivially adjusted to a broad variety of neuron models without requiring manual implementation work. Commonly, vector-Jacobian products (VJPs) are preferred over explicitly calculating Jacobians, since the Jacobians are usually large (see also Appendix \ref{app:param_grads}). However, in HYPR, we drastically reduced the dimensionality of these Jacobians $J_{f,i}^t$ and $J_{g,i}^t$ of functions $f$ and $g$ by dissecting the computation of scalar neuron input $I^t_i$ from the state-to-state transition function $f$. Hence, the Jacobians are computed w.r.t.~scalar input $I^t_i$ instead of high-dimensional parameter $\theta_{i}$ and hence are low-dimensional. This dissection allows the low-rank decomposed representation of parameter gradients $\delta^t$ described in Appendix \ref{app:param_grads}.

We can slice these Jacobians into partial derivatives $\frac{\partial \mathbf{s}_i^t}{\partial \mathbf{s}_i^{t-1}} \in \mathbb{R}^{k \times k}$, $\frac{\partial \mathbf{s}_i^t}{\partial I_i^{t}} \in \mathbb{R}^{k \times 1}$ and $\frac{\partial y_i^t}{\partial \mathbf{s}_i^{t}} \in \mathbb{R}^{k \times 1}$.
\section{Combining approximate backward and forward propagation}
\label{app:backward_ssm}
In this section we explain in detail how to obtain the backward formulation in Eq.~\eqref{eq:backward_ssm} from the forward-SSM formulation of the eligibility matrix from Eq.~\eqref{eq:elig_vector_ssm}. Consider a summative loss function $\mathcal{L} = \sum_t \mathcal{L}^t$. Then, APG $\apg$ (see Eq.~\eqref{eq:elig_vector_ssm2}) is given by 
\begin{align}
    \apg &= \frac{d \mathcal{L}^t}{d \mathbf{s}_i^t}\mathbf{e}_{i}^t \\
  &= \frac{d \mathcal{L}^t}{d \mathbf{s}_i^t}\left(\delta_{i}^t + \underbrace{A_i^t}_{\phi_i^{t:t}} \delta_{i}^{t-1} + \underbrace{A_i^t A_i^{t-1}}_{\phi_i^{t:t-1}} \delta_{i}^{t-2}  + \ldots + \underbrace{A_i^t \ldots  A_i^{2}}_{\phi_i^{t:2}} \delta_{i}^{1} + \underbrace{A_i^t \ldots  A_i^{1}}_{\phi_i^{t:1}} \mathbf{e}_{i}^{0}\right),
\end{align}
where in the second line, $\mathbf{e}_{i}^t$ is unrolled as in Eq.~\eqref{eq:elig_unrolled}.
The cumulative APG $[\apgnot]^{1:\lambda} = \sum_{t=1}^\lambda \apg$ is then given by the summation of all terms $\apg$ as
\begin{align}
\label{eq:cum_weight_upd_decomposed}
\sum_{t=1}^\lambda \apg
&=
\underbrace{\,\frac{d \mathcal{L}^1}{d \mathbf{s}_i^1}
\Bigl(\delta_{i}^{1}
+\phi_i^{1:1}\,\mathbf{e}_{i}^0\Bigr)}_{\apgnot^1} \\
&\qquad+\underbrace{\,\frac{d \mathcal{L}^2}{d \mathbf{s}_i^2}
\Bigl(\delta_{i}^{2}
+\phi_i^{2:2}\,\delta_{i}^{1}
+\phi_i^{2:1}\,\mathbf{e}_{i}^0\Bigr)}_{\apgnot^2} \nonumber\\
&\qquad+\cdots \nonumber\\
&\qquad+\underbrace{\frac{d \mathcal{L}^{\lambda-1}}{\delta_{i}^{\lambda-1}}
\Bigl(
\delta_{i}^{\lambda-1}
+\phi_i^{\lambda-1:\,\lambda-1}\,\delta_{i}^{\lambda-2}
+\phi_i^{\lambda-1:\,\lambda-2}\,\delta_{i}^{\lambda-3}
+\cdots
+\phi_i^{\lambda-1:\,1}\,\mathbf{e}_{i}^0
\Bigr)}_{\apgnot^{\lambda-1}} \nonumber\\
&\qquad+\underbrace{\frac{d \mathcal{L}^\lambda}{d \mathbf{s}_i^\lambda}
\Bigl(
\delta_{i}^{\lambda}
+\phi_i^{\lambda:\,\lambda}\,\delta_{i}^{\lambda-1}
+\phi_i^{\lambda:\,\lambda-1}\,\delta_{i}^{\lambda-2}
+\cdots
+\phi_i^{\lambda:\,1}\,\mathbf{e}_{i}^0
\Bigr)}_{\apgnot^\lambda}
\nonumber
\end{align}
where the term $\delta_{i}^{\lambda}$ appears once, the term $\delta_{i}^{\lambda-1}$ appears twice, and so on. A number of $\mathcal{O}(\lambda^2)$ cumulative state transition matrices $\phi^{r:s}$ with $r,s \in \{1,\ldots,\lambda\}, r\geq s$ would need to be calculated to parallelize this computation. Rearranging this equation by collecting all coefficients for $\delta_{i}^{1},\; \ldots, \; \delta_{i}^{\lambda}$ and $\mathbf{e}_{i}^0$ shows why the backward formulation is much more elegant to solve this problem:
\begin{align}
% \label{eq:reordered_cum_weight}
\sum_{t=1}^\lambda \Delta \theta_{i}^t
&= \ldots \nonumber\\[1ex]
&\quad+
\underbrace{\left(
  \frac{d \mathcal{L}^\lambda}{d \mathbf{s}_i^\lambda}
  \,A^\lambda A^{\lambda-1}
  +
  \frac{d \mathcal{L}^{\lambda-1}}{d \mathbf{s}_i^{\lambda-1}}
  A^{\lambda-1}
  +
  \frac{d \mathcal{L}^{\lambda-2}}{d \mathbf{s}_i^{\lambda-2}}
\right)}_{\mathbf{q}_i^{\lambda-2}}
 \delta_{i}^{\lambda-2}\nonumber\\
&\quad+
\underbrace{\left(
  \frac{d \mathcal{L}^\lambda}{d \mathbf{s}_i^\lambda}
  \,A^\lambda
  +
  \frac{d \mathcal{L}^{\lambda-1}}{d \mathbf{s}_i^{\lambda-1}}
\right)}_{\mathbf{q}_i^{\lambda-1}}
\delta_{i}^{\lambda-1} \nonumber\\
&\quad+ 
\underbrace{\left(
  \frac{d \mathcal{L}^\lambda}{d \mathbf{s}_i^\lambda}
\right)}_{\mathbf{q}_i^{\lambda}}
\delta_{i}^{\lambda} \nonumber\\
&= \mathbf{q}_i^\lambda \delta_{i}^{\lambda} + \mathbf{q}_i^{\lambda-1} \delta_{i}^{\lambda-1} +\ldots   + \mathbf{q}_i^1 \delta_{i}^{1} + \mathbf{q}_i^0 \mathbf{e}_{i}^{0}.
\label{eq:fully_resolved_weight_update}
\end{align}
This formulation is mathematically equivalent to Eq.~\eqref{eq:cum_weight_upd_decomposed}, but the individual APGs $\apg$ disappeared. This is possible, since we are only interested in their sum, not in the individual terms. The major advantage here is that vectors $\mathbf{q}_i^t \in \mathbb{R}^{1\times k}$ are much smaller than eligibility matrices $\mathbf{e}_{i}^t \in \mathbb{R}^{k \times d_{\theta}}$ and can be efficiently computed using the associative scan algorithm, as explained in Appendix \ref{app:backward_hypr}. Note, that for the implementation we replaced all $\delta_{i}^{t}$ by their low-rank representation from Appendix \ref{app:param_grads} which allows efficient parallel computation of Eq.~\eqref{eq:fully_resolved_weight_update} using Einstein summation.
\section{Computing $\mathbf{q}_i^t$ with the associative scan}
\label{app:backward_hypr}
Recall the linear SSM formulation from Eq.~\eqref{eq:q_recursive}:
\begin{equation}
    \mathbf{q}_i^t =  \mathbf{q}_i^{t+1}A_i^{t+1}  +  \ell^t_i
    \label{eq:q_recursive_app}
\end{equation}
with $\ell^t_i=\frac{d \mathcal{L}^t}{d \mathbf{s}_i^t}$. We can efficiently obtain the sequence $[\mathbf{q}_i^\lambda,\;\ldots,\;\mathbf{q}_i^0] = [\ell^\lambda_i,\;\; \ell^{\lambda-1}_i + \ell^\lambda_i A_i^{\lambda},\;\;  \ell^{\lambda-2}_i + \ell^{\lambda-1}_i A_i^{\lambda-1} + \ell^\lambda_i  A_i^{\lambda} A_i^{\lambda-1}]$ using the associative scan \cite{BlellochTR90}, as described in \cite{smithSimplifiedStateSpace2023a}: Given any sequence $[a,b,c,\ldots]$ of length $\lambda$ and a binary associative operator $\bullet$ which satisfies $(a\bullet b) \bullet c = a\bullet (b \bullet c)$, the sequence $[a,\;\; a\bullet b,\;\; a\bullet b \bullet c,\;\; \ldots]$ can be computed in $\mathcal{O}(\log \lambda)$ time complexity, given sufficient parallel processors are available.

We define tuples $p^t_i\myeq(A_i^{t+1},\ell^t_i)$ with $A_i^{\lambda+1}= \mathbf{I}$ as identity matrix, $\ell^0_i=0$ and binary associative operator $(a_1,a_2)\bullet (b_1, b_2) \myeq (b_1 \times a_1,\; b_1 a_2 + b_2)$, where subscripts $1$ and $2$ refer to the first and second tuple element respectively. Applying the associative scan using this binary associative operator on the sequence $[p^\lambda_i,\;\ldots,\;p^0_i]$ results in the sequence $[r^\lambda_i,\;\ldots,\;r^0_i]$ with
\begin{align}
    r^\lambda &= p^\lambda &= (A^{\lambda+1},\ell^\lambda) \\
    % p^{\lambda-1} = (A^{\lambda},\ell^(\lambda-1))
    r^{\lambda-1} &= p^\lambda \bullet p^{\lambda-1} &= (A^{\lambda}  A^{\lambda+1},\; A^{\lambda} \ell^{\lambda} + \ell^{\lambda-1}) \\
    r^{\lambda-2} &= p^\lambda \bullet p^{\lambda-1} \bullet p^{\lambda-2} &= (\underbrace{A^{\lambda-1} A^{\lambda} A^{\lambda+1}}_{\phi^{\lambda-2:\lambda+1}},\; \underbrace{A^{\lambda-1}A^{\lambda} \ell^{\lambda} + A^{\lambda-1}\ell^{\lambda-1} + \ell^{\lambda-2}}_{\mathbf{q}^{\lambda-2}}) \\
    \vdots
\end{align}
where we omitted neuron index $i$ for readability. From each resulting tuple $r^t=(\phi^{t:\lambda+1}, \mathbf{q}^t)$ we can extract  $\mathbf{q}^t$ as the second tuple element.
\section{Schematic illustration of HYPR}
\label{app:schema_illustration}
\begin{figure}[!ht]
    \centering
    \includegraphics[width=\linewidth]{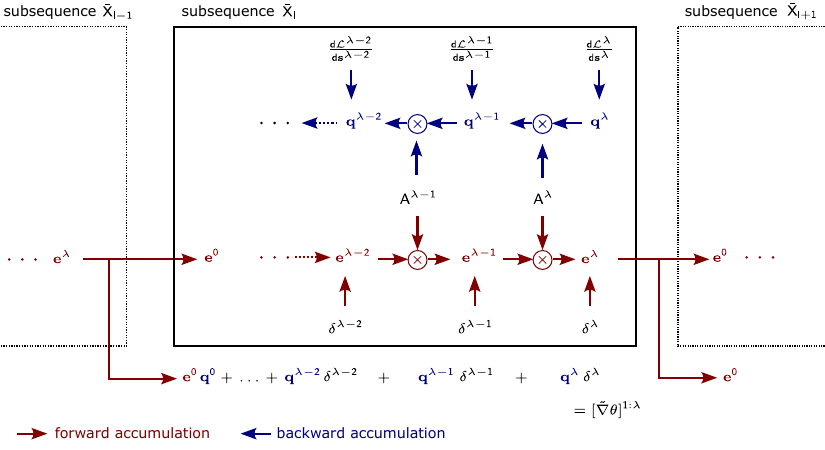}
    \caption{Schematic illustration of the combination of forward-accumulation (Eq.~\eqref{eq:elig_vector_ssm}) and backward-accumulation (Eq.~\eqref{eq:q_recursive}) in HYPR.}
    \label{fig:schema}
\end{figure}

Fig.~\ref{fig:schema} shows a schematic illustration of the forward and backward accumulation in HYPR.
\section{Extension of HYPR to multi-layer networks}
\label{app:multi_layer}
%\checked{}
HYPR can be applied to multi-layered networks. This can trivially be achieved by back-propagating the loss $\mathcal{L}^t$ obtained in time step $t$ through layers to obtain $\fraci{d \mathcal{L}^t}{d \mathbf{s}_{l,i}^t}$, where $\mathbf{s}_{l,i}$ denotes the state of neuron $i$ in layer $l$ at time step $t$, which can directly be plugged into the computation of APG $\apgl$ (see Eq.~\eqref{eq:elig_vector_ssm2}) of layer $l$. However, it has to be noted that with this approach some gradient pathways are disregarded compared to BPTT: The exact gradient $\fraci{d \mathcal{L}^t}{d \mathbf{s}_{l-1,i}^r}$ with $r<t$ of layer $l-1$ can be expressed as 
\begin{equation}
    \frac{d \mathcal{L}^t}{d \mathbf{s}_{l-1,i}^r} = 
    \sum_k\frac{d \mathcal{L}^t}{d \mathbf{s}_{l,k}^r}\frac{\partial \mathbf{s}_{l,k}^r}{\partial \mathbf{s}_{l-1,i}^r}
    + 
    \sum_k
    \frac{d \mathcal{L}^t}{d \mathbf{s}_{l-1,k}^{r+1}}
    \frac{\partial \mathbf{s}_{l-1,k}^{r+1}}{\partial \mathbf{s}_{l-1,i}^r}.
\end{equation}
BPTT accounts for all gradient pathways involved. In contrast, the APGs $\apgl$ used in HYPR disregard the terms 
$\frac{d \mathcal{L}^t}{d \mathbf{s}_{l,k}^r}
\frac{\partial \mathbf{s}_{l,k}^r}{\partial \mathbf{s}_{l-1,i}^r}$ for $r<t$. 
These gradient terms capture how state $\mathbf{s}_{l-1,i}^r$ indirectly influences loss $\mathcal{L}^t$ through states ($\mathbf{s}_{l,k}^r,\ldots,\mathbf{s}_{l,k}^{t-1}$) of successive layer $l$. The resulting approximation $[\fraci{d \mathcal{L}^t}{d \mathbf{s}_{l-1,i}^r}]_\text{local}$ of HYPR is given by
\begin{align}
    \left[\frac{d \mathcal{L}^t}{d \mathbf{s}_{l-1,i}^r}\right]_\text{local} &= 
    \left\{
\begin{array}{ll}
\frac{\partial \mathbf{s}_{l-1,k}^{r+1}}{\partial \mathbf{s}_{l-1,i}^r}\left[\frac{d \mathcal{L}^t}{d \mathbf{s}_{l-1,i}^{r+1}}\right]_\text{local}   & \text{if} \quad  r<t \\
\frac{d \mathcal{L}^t}{d \mathbf{s}_{l-1,i}^r} & \text{if} \quad r=t \\
\end{array}
\right. 
\end{align}
The terms $[\fraci{d \mathcal{L}^t}{d \mathbf{s}_{l-1,i}^r}]_\text{local}$ are never explicitly computed in HYPR, we provide it merely to demonstrate which terms are ignored by HYPR in multi-layered networks. For the multi-layer networks, we compute the spatially back-propagated loss gradients $\fraci{d \mathcal{L}^t}{d \mathbf{s}_{l-1,i}^t}$ and use them in Eq.~\eqref{eq:backward_ssm}. The approximation of the multi-layer case discussed here is inherent to online algorithms and has previously been discussed in \cite{zucchetOnlineLearningLongrange2023} and \cite{onlinetrainingthroughtime}.
\section{Performance of multi-layer networks trained with HYPR}
\label{app:multi_layer_results}
Tab.~\ref{tab:multi_layer_hypr} shows the performance of SRNNs consisting of $1-3$ layers of SE-adLIF neurons in the ECG task, trained with HYPR. All other hyperparameters were equal to the ones reported in Appendix \ref{app:hyperparams}. Despite the approximations discussed in Appendix \ref{app:multi_layer}, the test accuracy improves when more layers are added to the network.
\begin{table}[ht]
\centering
\caption{Performance of a multi-layer SE-adLIF network trained with HYPR. We report mean $\pm$ std.~dev.~over $5$ random seeds which randomized initialization and train/validation split.}
\label{tab:multi_layer_hypr}
\vspace{0.1cm}
% SHD dataset
\begin{tabular}{ll}
\toprule
 \textbf{No.~of layers} & \textbf{Test Acc.[\%]} \\
\midrule
1 & 80.17 $\pm$ 0.34 \\
2 & 83.00 $\pm$ 0.78 \\
3 & 84.35 $\pm$ 0.35 \\
\bottomrule
\end{tabular}
\end{table}
\section{Details of datasets and preprocessing}
\label{app:details_datasets}
For all datasets except cue and Pathfinder-E, a train/test split is predefined. We further split the training set into a smaller training set and a validation set with $90\%$ and $10\%$ relative size respectively and perform model selection on the validation set. For Pathfinder-E and cue, we split training/validation/test as $80/10/10\%$.
\subsection{Details of the cue task}
The cue task is a binary classification task designed to test a model's ability to remember a class cue over an extended delay and indicate its identity during a recall input. Each input pattern $\mathbf{X} \in \{0,1\}^{T \times D}$ in this task consists of spike trains across $D = 15$ neurons and a total duration of $T = 2T_{\text{pat}} + T_{\text{delay}}$ time steps, where $T_{\text{pat}}$ is the pattern length and $T_{\text{delay}}$ is the delay length. The task is to classify each pattern into class A or B according to the activation pattern presented before the delay period. 

\begin{figure}[!t]
    \centering
    \includegraphics[width=\linewidth]{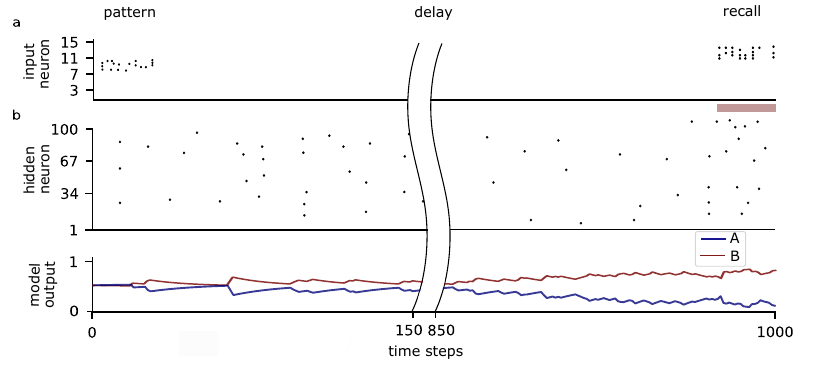}
     \caption{The cue task. \textbf{a} One input sample from class B with a length of $T=1000$ time steps. Red bar below input neurons indicates the time when a target signal is available to the network. \textbf{b} Hidden layer activity (top, subsample of $100$ neurons) and model output (bottom) of a single hidden layer RSNN trained on this task. Red (blue) lines indicate the network's predicted probability for class A (B).}
    \label{fig:cue_dataset_B}
\end{figure}

To generate input patterns for this task, we employed a three-phase process in which we first randomly assign a target class, then generate class-specific input patterns, followed by a delay period and a class-independent recall cue signal. An example input pattern is shown in Fig.~\ref{fig:cue_dataset_B}a. In detail, this procedure is as follows:
For each pattern, a binary target class $c \in \{A, B\}$ is randomly assigned with equal probability. This target determines which set of neurons will be active during the initial pattern phase. The pattern phase has a duration of $T_{\text{pat}}$ time steps. If the target is class A, input neurons $1$ to $5$ are active with probability $p_A = 0.5$ at each time step, generating a binary pattern where each spike $x_{t,i} \sim \text{Bernoulli}(p_A)$. If the target is class B, input neurons $6$ to $10$ are activated according to the same mechanism. All other neurons remain silent during this phase.
Following the pattern phase, a delay period of length $T_{\text{delay}}$ time steps occurs, during which all neurons remain silent. This creates a temporal gap between the presentation of the class-specific pattern and the recall cue.

The final phase is the recall phase, which also lasts for $T_{\text{pat}}$ time steps. During this phase, input neurons $11$ to $15$ are active according to the same process as used in the pattern phase. 
Each target is represented as a one-hot encoded vector $\mathbf{y} \in \{0, 1\}^2$ and presented to the network only during the recall period.

Fig.~\ref{fig:cue_dataset_B}b shows the hidden layer spiking activity and output layer softmax probabilities of an RSNN that was trained on this task.
The correct class probability rises in response to the recall cue. 

For our experiments, we used $T_{\text{pat}} = 20$ time steps and created multiple variants of the data set with different delay lengths $T_{\text{delay}} \in \{1k, 2k, 5k, 10k, 16k\}$ time steps to test the model's capacity to maintain information over increasingly longer temporal gaps. For each variant, we generated a total of 256 samples, with an equal number of samples for each class. After generating an entire dataset, samples were randomly shuffled and split into a training set ($80\%$) and test set ($20\%$). 
\subsection{Details of the benchmark datasets}
\label{app:details_benchmark}
\textbf{\textsc{SHD}:}
The Spiking Heidelberg Digits (SHD) dataset is an audio-based classification dataset for benchmarking SNNs.
It consists of 20 classes, corresponding to the spoken digits 0 to 9 in German and English, where the audio is converted into spike trains based on a detailed cochlea model~\citep{cramerHeidelbergSpikingData2022}.
We use the version from the Tonic python library (version 1.5.1) for neuromorphic datasets\footnote{\url{https://tonic.readthedocs.io/en/latest/generated/tonic.datasets.SHD.html}}, which is publicly available for research (Creative Commons Attribution 4.0 International License). We reduced the dimension from $700$ to $140$ channels by sum pooling and summed the spikes over $4$ ms time windows. The same preprocessing has been applied in \citep{baronig2025advancingspatiotemporalprocessingspiking}. 

\textbf{\textsc{ECG}:}
The electrocardiogram (ECG) dataset involves signals with six distinct characteristic waveforms, whose shape and duration is informative of functioning of the cardiovascular system. 
The task consists of recognition of the $6$ classes per time-step.
We use the dataset\footnote{\url{https://github.com/byin-cwi/Efficient-spiking-networks/tree/main/data}} version preprocessed and utilized by the ALIF paper~\cite{yin2021accurateefficienttimedomainclassification}, which is referenced to the original publicly available QT Database from PhysioNet\footnote{\url{https://physionet.org/content/qtdb/1.0.0/}} \citep{laguna1997database} under the Open Data Commons Attribution License v1.0.

\textbf{\textsc{sMNIST}:}
In this task, the 28x28 dimensional grayscale images of the MNIST dataset~\citep{lecun1998mnist} are presented in a pixel-by-pixel sequence of length 784, and the task is to decide to which one of the 10 handwritten digits is present in the current image.
This sequential MNIST (sMNIST) task formulation was initially introduced in~\citep{bellec2018long}.
We provide the input to the networks after normalization of pixel values to be between $[0,1]$. 
We access to this publicly available dataset via the Tensorflow Datasets library\footnote{\url{https://www.tensorflow.org/datasets/catalog/mnist}}.

\textbf{\textsc{sCIFAR}:}
In this task, the 32x32 dimensional colored images of the CIFAR-10 dataset~\citep{Krizhevsky:2009} are presented in a pixel-by-pixel sequence of length 1024, and the task is to decide to which one of the 10 categories the current image belongs to.
We provide each input in RGB, thus the inputs contain three channels. 
Inputs are provided after normalization of pixel values to be between $[0,1]$. 
We access to this publicly available dataset via the Tensorflow Datasets library\footnote{\url{https://www.tensorflow.org/datasets/catalog/cifar10}}.
This task is also one of the long-range arena benchmark tasks~\citep{tay2020long}, under category ``Image''.

\textbf{\textsc{Pathfinder-E}:}
It is one of the long-range arena benchmark tasks~\citep{tay2020long}, where a 32x32 grayscale image of line drawings is presented in a pixel-by-pixel sequence of length 1024, and the task is to decide whether a starting point is connected by a line with an end point (i.e., binary classification).
Our dataset implementations are based on the code repository (V2 release)\footnote{\url{https://github.com/google-research/long-range-arena/}}
and the publicly available long range arena repository\footnote{\url{https://storage.googleapis.com/long-range-arena/lra_release.gz}}.
We use the easier variant of this task (-E), with the difficulty level indicated as ``baseline'' in the dataset, as opposed to the harder variant with additional distracting contours present in the image.
\section{Details of neuron models}
\label{app:neuron_models}
As shown in Table~\ref{tab:neuron_models}, each neuron model can be represented by its state vector $\mathbf{s}$, update function $f$, and output function $g$. In all neuron models, output function $g$ is the Heaviside step function $\Theta$.
\begin{table}[ht]
    \centering
    \caption{$f$ and $g$ functions for different neuron models. In the ALIF and SE-adLIF models, the variable $u^t_i$ is set to $0$ (hard reset) if the neuron spiked in the previous time step (not shown).}
    \label{tab:neuron_models}\vspace{0.1cm}
    \begin{tabular}{c c c c}
        \toprule
            Model & $\mathbf{s}$ & $f$ & $g$ \\
            \midrule
            \vspace{0.5cm}
            BRF &$\begin{bmatrix} u_i^t \\ v_i^t \\ q_i^t \end{bmatrix}$ &$\begin{bmatrix} 
 u_i^{t-1} +\Delta t(b_i^t u_i^{t-1} - \omega v_i^{t-1} +  I_i^t) \\
v_i^{t-1} + \Delta t(\omega u_i^{t-1} + b_i^t v_i^{t-1}) \\
\gamma q_i^{t-1} + z_i^t
\end{bmatrix}$ & $y_i^t = g(\mathbf{s}^t_i) = \Theta(u_i^t - \theta - q_i^{t-1})$ \\
            \vspace{0.5cm}
            SE-adLIF &$\begin{bmatrix} u_i^t \\ w_i^t \end{bmatrix}$  & $\begin{bmatrix} 
\alpha u_i^{t-1} + (1-\alpha)(I_i^t - w_i^{t-1}) \\
\beta w_i^{t-1} + (1-\beta)(a u_i^{t} + b y_i^{t-1})
\end{bmatrix}$&$y_i^t = g(\mathbf{s}^t_i) = \Theta(u_i^t - \theta)$ \\
            ALIF & $\begin{bmatrix} u_i^t \\ a_i^t \end{bmatrix}$&$\begin{bmatrix} 
\alpha u_i^{t-1} + (1-\alpha)I_i^t - A_i^t y_i^{t-1} \\
\rho a_i^{t-1} + (1-\rho)y_i^{t-1}
\end{bmatrix} $ & $y_i^t = g(\mathbf{s}^t_i) = \Theta(u_i^t - A_i^t)$ \\
            \bottomrule
    \end{tabular}
\end{table}
\subsection{Balanced Resonate-and-Fire (BRF)}
The dynamics of the BRF~\citep{higuchi2024balanced} neuron model are given by:
\begin{align}
b^t &= p_{\omega} - b_{\text{offset}} - q^{t-1} \\
u^t &= u^{t-1} + \Delta t \cdot (b^t \cdot u^{t-1} - \omega \cdot v^{t-1} + I^t) \\
v^t &= v^{t-1} + \Delta t \cdot (\omega \cdot u^{t-1} + b^t \cdot v^{t-1}) \\
z^t &= \Theta( u^t - \theta - q^{t-1}) \\
q^t &= \alpha \cdot q^{t-1} + z^t
\end{align}
\noindent where $\theta$ is the base firing threshold, $\omega$ is the angular frequency parameter controlling oscillations, $p_{\omega} = \frac{-1 + \sqrt{1 - (\Delta t \cdot \omega)^2}}{\Delta t}$ is the divergence boundary, $b_{\text{offset}}$ is the dampening parameter, $q^t$ is an adaptation variable with $\alpha=0.9$ as adaptive decay factor, and $\Delta t$ is the simulation time step. We use $\Delta t = 0.01$ in all experiments. The BRF neuron generates a spike when the membrane potential $u^t$ exceeds the adaptive threshold $\theta + q^{t-1}$. In the BRF model the membrane potential is not reset after a spike. Instead, the adaptation variable $q^t$ increases by the spike output $z^t \in {0,1}$ and affects both the threshold for future spikes and the dampening term $b^t$ in the subsequent dynamics. We train $b_{\text{offset}}$ and $\omega$ together with the network weights.
We used the same weight initialization as in \citep{higuchi2024balanced}. 
\subsection{SE discretized adaptive Leaky Integrate and Fire (SE-adLIF) }
The temporal dynamics of the Symplectic-Euler(SE) discretized SE-adLIF~\citep{baronig2025advancingspatiotemporalprocessingspiking} neuron are given by
\begin{align}
% Discretized equations with time step dt
\hat{u}^t &= \alpha u^{t-1} + (1-\alpha)(I^t-w^{t-1}) \\
z^t &= \Theta( \hat{u}^t - \theta) \\
u^t &= \hat{u}^t(1-z^t) \\
w^t &= \beta w^{t-1} + (1-\beta)(au^t + bz^t) 
\end{align}
\noindent where $\tau_u$ and $\tau_w$ are the membrane potential and adaptation time constants, and $a = \rho\hat{a}, b= \rho\hat{b}$ are adaptation parameters. For all experiments, we keep $\rho=120$ fixed and initialize $\hat{a},\hat{b} \sim \mathcal{U}(0,1)$, which are trained together with other parameters. During training we clip both $\hat{a}$ and $\hat{b}$ in the range $[0,1]$. Further, we employ the same time constant interpolation as in \citep{baronig2025advancingspatiotemporalprocessingspiking} with the ranges $\tau_u\in[5,25]$ and $\tau_w\in[60,300]$ for all experiments.
We used the same weight initialization as in ~\citep{baronig2025advancingspatiotemporalprocessingspiking}.
\subsection{Adaptive Leaky Integrate and Fire (ALIF) }
The dynamics of the ALIF~\citep{yin2021accurateefficienttimedomainclassification} model are given by 
\begin{align}
a^t &= \rho \cdot a^{t-1} + (1-\rho) \cdot z^{t-1} \\
A^t &= b_{j0} + \beta \cdot a^t \\
u^t &= \alpha \cdot u^{t-1} + (1-\alpha) \cdot I^t - A^t \cdot z^{t-1} \\
z^t &= \Theta( u^t - A^t)
\end{align}
\noindent Where $\alpha = e^{-\Delta t/\tau_u}$ is the membrane potential decay factor, $\rho = e^{-\Delta t/\tau_a}$ is the adaptation decay factor, where $\tau_u$ is the membrane potential time constant, $\tau_a$ is the adaptation time constant. $\beta$ is the adaptation strength coefficient, and $A^t$ is the adaptive threshold. 
The ALIF neuron implements an adaptive threshold mechanism combined with a spike-triggered reset. The adaptation variable $a^t$ tracks the neuron's recent spiking history, increasing with each spike $z^{t-1}$. This creates a dynamic threshold $A^t = b_{j0} + \beta \cdot a^t$ that rises after spiking activity and implementing spike-frequency adaptation. After a spike, membrane potential $u^t$ is reset to 0.
\subsection{Leaky Integrator (LI)}
The dynamics of the LI neurons in the output layer are given by
\begin{align}
u^t &= \alpha \cdot u^{t-1} + (1-\alpha) \cdot I^t
\end{align}
\noindent where $\alpha = e^{-\Delta t/\tau_u}$ is the decay factor with time constant $\tau_u$.

\section{Training details and hyperparameters}
\label{app:training_and_hyperparams}
\subsection{Training details}
\textbf{Surrogate gradients:} We employed surrogate gradient functions that approximate the derivative of $\Theta(x)$. In our experiments, we utilize two common surrogate gradients. The first is the SLAYER~\citep{shrestha2018slayer} surrogate gradient, defined as $\frac{d\Theta(x)}{dx} \approx \alpha  c   e^{-\alpha|x|},$ where $\alpha=5$ controls the sharpness of the exponential curve and $c=0.2$ adjusts the amplitude of the gradient. The second approach is the double Gaussian~\cite{yin2021accurateefficienttimedomainclassification} surrogate gradient, given by $\frac{d\Theta(x)}{dx} \approx \gamma \left[(1+p) \cdot \text{G}(x; 0, \sigma_1) - 2p \cdot \text{G}(x; 0, \sigma_2) \right],$ where $\text{G}(x; \mu, \sigma)$ represents the Gaussian probability density function, $\sigma_1 = 0.5$ and $\sigma_2 = 6\sigma_1$ control the width of the Gaussian curves, $p=0.15$ adjusts the relative weight between the two Gaussians, and $\gamma=0.5$ is an overall scaling factor.

\textbf{Optimizer:} We used the ADAM~\citep{kingma2017adam} optimization algorithm with $\beta_1=0.9$, $\beta_2=0.999$ and $\epsilon = 1\text{e}-8$ but with different learning rates (see Appendix \ref{app:hyperparams}) for all experiments. For HYPR, we accumulated APGs over the entire sequence before applying the weight updates. We applied gradient clipping for all experiments: if the gradient norm exceeded a certain magnitude, we rescaled it to $m_\text{grad}$. The values of $m_\text{grad}$ are discussed in Appendix \ref{app:hyperparams}.

\textbf{Learning rate schedule:} We explored three different learning rate scheduling methods: constant, linear and cosine. In the constant scheduler, the learning rate was held constant throughout all training epochs. In the linear scheduler, we decayed it linearly from an initial learning rate $\eta^{\text{init}}$ to $0$ at the final epoch. In the cosine decay scheduler~\citep{loshchilov2022sgdr}, we define the learning rate $\eta^k$ at the $k$-the epoch as $\eta^k = \eta^{\text{init}}\cdot[0.5\cdot(1+\cos{(\pi\cdot k/{\text{\footnotesize\#epochs}})})]$.

\textbf{Loss functions:} Our network architecture uses leaky integrator neurons in the output layer that match the number $C$ of classes of the corresponding task. At each time step $t$, the network output $\bar{\mathbf{y}}^t \in \mathbb{R}^C$ was given by the vector of membrane potentials $\mathbf{u}^t \in \mathbb{R}^C$. Depending on the task and the training algorithm (BPTT or HYPR), we used different functions to compute a loss from the series of outputs $[\bar{\mathbf{y}}^1,\;\ldots,\;\bar{\mathbf{y}}^T]$.
The \textit{sum-of-softmax} loss was given by
\begin{equation}
    \mathcal{L} = \text{CE}\left(\text{softmax}\biggl(\sum_{t=t_0+1}^T \text{softmax}(\bar{\mathbf{y}}^t)\biggr), \mathbf{y}^*\right),
\end{equation}
where CE denotes the cross entropy loss, softmax is the softmax function, $\mathbf{y}^* \in \mathbb{R}^C$ a one-hot encoded target vector and $t_0$ defines the time step up to which the network output is ignored. $t_0$ is a hyperparameter and varies between different tasks and models and is shown in Appendix \ref{app:hyperparams}. The sum-of-softmax is not compatible with HYPR since no per-timestep loss $\mathcal{L}^t$ can be obtained at time step $t$ without back-propagating the loss through time. Therefore, we used a summative \textit{per-timestep loss}, given by
\begin{equation}
    \mathcal{L} = \frac{1}{T-t_0}\sum_{t=t_0+1}^T \mathcal{L}^t = \frac{1}{T-t_0}\sum_{t=t_0+1}^T\text{CE}\left(\text{softmax}\biggl(\sum_{t=t_0+1}^T \bar{\mathbf{y}}^t\biggr), \mathbf{y}^{*t}\right),
\end{equation}
where $\mathbf{y}^{*t}$ can be a per-timestep target (for example in the ECG task) or is the same for every time step (for example in SHD).

\textbf{Class prediction:}
To compute the accuracy, we obtained a class prediction $\hat{y}$ from $[\bar{\mathbf{y}}^1,\;\ldots,\;\bar{\mathbf{y}}^T]$ via three different methods:
\begin{align}
    \hat{y}\; &= \text{argmax}\biggl(\sum_{t=t_0+1}^T \text{softmax}(\bar{\mathbf{y}}^t)\biggr) & \text{(sum of softmax)} \\
    \hat{y}\; &= \text{argmax}\biggl(\sum_{t=t_0+1}^T \bar{\mathbf{y}}^t\biggr) & \text{(mean over time)} \\
    \hat{y}^t &= \text{argmax}(\bar{\mathbf{y}}^t) & \text{(per timestep)}.
\end{align}
The method of choice for each task and model is shown in Appendix \ref{app:hyperparams}.  

\textbf{Frameworks:}
We used the open source Python frameworks Jax \citep{jax2018github}, for calculation and automatic differentiation, Hydra\footnote{\url{https://hydra.cc}}, for configuring experiments, and Aim\footnote{\url{https://aimstack.io}} for experiment tracking.
\subsection{Hyperparameters and Tuning}
\label{app:hyperparams}
The hyperparameters for the BRF model trained on the cue task to obtain the plots in Fig.~\ref{fig:profiling} are shown in Tab.~\ref{tab:cue_brf_profiling}. Tables~\ref{tab:model_comparison_shd},~\ref{tab:model_comparison_ecg},~\ref{tab:model_comparison_smnist}, and ~\ref{tab:model_comparison_lra} show the hyperparameters used for training the neuron models on each benchmark dataset. For each model and task, we manually tuned the number of neurons, batch size, number of layers, surrogate gradient and number of ignored time steps. Since it is impossible to run all hyperparameter configurations, we oriented the search initially on the configurations from the authors of the original model and tuned the hyperparameters with educated guesses and small grid searches. Increasing the number of parameters did not always result in better test accuracy due to overfitting, hence the number of neurons (and therefore the number of parameters) varies between different neuron models and tasks. 
\begin{table}[t!]
\centering
\caption{List of hyperparameters corresponding to the \textbf{cue} dataset simulations with the BRF model from Fig.~\ref{fig:profiling}. Hyperparameters that were different for HYPR are denoted in parentheses. For this experiment we used $\alpha=1$ and $c=0.2$ for the SLAYER surrogate gradient. $m_\text{grad}$ is the gradient clipping magnitude.}
\label{tab:cue_brf_profiling}
\vspace{0.1cm}
% SHD dataset
\begin{tabular*}{0.6\textwidth}{l@{\extracolsep{\fill}}l}
\toprule
Hyperparameter & Value \\
\midrule 
learning rate (lr) & 0.1 (0.01) \\
lr scheduler & constant  \\
nb.~layers &1  \\
epochs & 200 \\
%dropout &0.0 & 0.0&0.1 \\
batch size & 128 \\
nb.~neurons & 1024\\
\multirow{3}{*}{parameter initialization} 
  & $\omega$: $\mathcal{U}(0.01,10)$ \\
  & b: $\mathcal{U}(1e-9,1e-4)$  \\
  & $\tau_{out}$: $\mathcal{U}(15,25)$ \\
loss aggregation & per timestep CE \\
prediction mode & mean over sequence \\
surrogate gradient & SLAYER   \\
grad.~clipping $m_\text{grad}$ & 10\\
% slayer params &$\alpha$:5 c:0.4 &$\alpha$:5 c:0.4 & $\alpha$:5 c:0.4\\
\bottomrule
\end{tabular*}
\end{table}
\begin{table}[t!]
\centering
\caption{List of hyperparameters corresponding to the \textbf{\textsc{SHD}} dataset simulations. Hyperparameters that were different for HYPR are denoted in parentheses. $t_0$ defines the time step up to which the network output is ignored for the loss computation and class prediction. $m_\text{grad}$ is the gradient clipping magnitude.}
\label{tab:model_comparison_shd}
\vspace{0.1cm}
% SHD dataset
\begin{tabular*}{\textwidth}{l@{\extracolsep{\fill}}lll}
\toprule
& \textbf{BRF} & \textbf{SE-adLIF} & \textbf{ALIF} \\
\midrule
learning rate (lr) &0.003 & 0.01& 0.01 \\
lr scheduler &linear & constant&constant \\
nb.~layers &1  &2 & 2\\
epochs &20&300 &300 \\
%dropout &0.0 & 0.0&0.1 \\
batch size &256 & 256& 256\\
nb.~neurons & 256& 360& 256\\
\multirow{3}{*}{parameter initialization} 
  & $\omega$: $\mathcal{U}(5,10)$ & $\tau_u$: $\mathcal{U}(5,25)$ & $\tau_u$: $\mathcal{N}(20,5)$ \\
  & b: $\mathcal{U}(2,3)$ & $\tau_w$: $\mathcal{U}(60,300)$ & $\tau_a$: $\mathcal{N}(150,10)$ \\
  & $\tau_{out}$: $\mathcal{U}(15,25)$ & $\tau_{out}$: $\mathcal{U}(15,15)$ & $\tau_{out}$: $\mathcal{U}(5,20)$ \\
loss aggregation & per timestep CE & sum of softmax & sum of softmax \\
prediction mode & mean over sequence & sum of softmax & sum of softmax \\
$t_0$&200 &50 &50 \\
surrogate gradient & DG  & SLAYER &SLAYER \\
grad.~clipping $m_\text{grad}$ & -- (1.0) & 10 & 10 \\
% slayer params &$\alpha$:5 c:0.4 &$\alpha$:5 c:0.4 & $\alpha$:5 c:0.4\\
\bottomrule
\end{tabular*}
\end{table}
\begin{table}[t!]
\centering
\caption{List of hyperparameters corresponding to the \textbf{\textsc{ECG}} dataset simulations. Hyperparameters that were different for HYPR are denoted in parentheses. $t_0$ defines the time step up to which the network output is ignored for the loss computation and class prediction. $m_\text{grad}$ is the gradient clipping magnitude.}
\label{tab:model_comparison_ecg}
\vspace{0.1cm}
% ECG dataset
\begin{tabular*}{\textwidth}{l@{\extracolsep{\fill}}lll}
\toprule
& \textbf{BRF} & \textbf{SE-adLIF} & \textbf{ALIF} \\
\midrule
learning rate (lr) &0.1 &0.01(0.005) &0.05(0.01) \\
lr scheduler &linear &constant & constant\\
nb.~layers &1 & 2&2 \\
epochs &300 &300 &300 \\
%dropout & 0.0& 0.15&0.0 \\
batch size &16 & 64(32)&16 \\
nb.~neurons & 36& 64&36 \\
\multirow{3}{*}{parameter initialization} 
  & $\omega$: $\mathcal{U}(3,5)$ & $\tau_u$: $\mathcal{U}(5,25)$ & $\tau_u$: $\mathcal{N}(20,0)$ \\
  & b: $\mathcal{U}(0.1,1.0)$ & $\tau_w$: $\mathcal{U}(60,300)$ & $\tau_a$: $\mathcal{N}(100,0)$ \\
  & $\tau_{out}$: $\mathcal{U}(15,25)$ & $\tau_{out}$: $\mathcal{U}(3,3)$ & $\tau_{out}$: $\mathcal{U}(5,5)$ \\
loss aggregation & per timestep CE & per timestep CE & per timestep CE  \\
prediction mode & per timestep & per timestep & per timestep \\
$t_0$ &0 & 50 & 50 \\
surrogate gradient & DG & SLAYER& SLAYER \\
grad.~clipping $m_\text{grad}$ & -- & 10 & 10 \\
% slayer params & & & \\
\bottomrule
\end{tabular*}
\end{table}
\begin{table}[t!]
\centering
\caption{List of hyperparameters corresponding to the \textbf{\textsc{sMNIST}} dataset simulations. Hyperparameters that were different for HYPR are denoted in parentheses. $t_0$ defines the time step up to which the network output is ignored for the loss computation and class prediction. $m_\text{grad}$ is the gradient clipping magnitude.}
\label{tab:model_comparison_smnist}
\vspace{0.1cm}
% SMNIST dataset
\begin{tabular*}{\textwidth}{l@{\extracolsep{\fill}}lll}
\toprule
& \textbf{BRF} & \textbf{SE-adLIF} & \textbf{ALIF} \\
\midrule
learning rate (lr) &0.1(0.01)& 0.01(0.001)& 0.008(0.003) \\
lr scheduler &linear & constant&constant \\
nb.~layers & 1& 2&2 \\
epochs & 300&300 &300 \\
%dropout &0.0 &0.0 &0.0 \\
batch size & 256 &512 & 256\\
nb.~neurons & 256& 360&256 \\
\multirow{3}{*}{parameter initialization} 
  & $\omega$: $\mathcal{U}(15,50)$ & $\tau_u$: $\mathcal{U}(5,25)$ & $\tau_u$: $\mathcal{N}(20,5)$ \\
  & b: $\mathcal{U}(0.1,1.0)$ & $\tau_w$: $\mathcal{U}(60,300)$ & $\tau_a$: $\mathcal{N}(200,25)$ \\
  & $\tau_{out}$: $\mathcal{U}(15,25)$ & $\tau_{out}$: $\mathcal{U}(15,15)$ & $\tau_{out}$: $\mathcal{U}(20,20)$ \\
loss aggregation & per timestep CE & sum of softmax & sum of softmax \\
prediction mode & mean over sequence &sum of softmax &sum of softmax \\
$t_0$& 500& 500 &0 \\
surrogate gradient & DG & SLAYER& SLAYER \\
grad.~clipping $m_\text{grad}$ & 1 & 10 & 10\\

\bottomrule
\end{tabular*}
\end{table}
\begin{table}[t!]
\centering
\caption{List of hyperparameters corresponding to the \textbf{\textsc{sCIFAR}} and \textbf{\textsc{Pathfinder-e}} dataset simulations with the \textbf{BRF} neuron model~\cite{higuchi2024balanced}. Hyperparameters that were different for HYPR are denoted in parentheses. $t_0$ defines the time step up to which the network output is ignored for the loss computation and class prediction. $m_\text{grad}$ is the gradient clipping magnitude.}
\label{tab:model_comparison_lra}
\vspace{0.1cm}
% sCIFAR PATHFINDER dataset
\begin{tabular*}{\textwidth}{l@{\extracolsep{\fill}}lll}
\toprule
& \textbf{\textsc{sCIFAR}} & \textbf{\textsc{Pathfinder-e}}  \\
\midrule
learning rate (lr) & 0.02 & 0.002 \\
lr scheduler & cosine & cosine \\
nb.~layers & 1 & 1 \\
epochs & 300 & 300 \\
%dropout & &  \\
batch size & 128 & 128 \\
nb.~neurons & 1024 & 1024 \\
\multirow{3}{*}{parameter initialization} 
  & $\omega$: $\mathcal{U}(15,50)$ & $\omega$: $\mathcal{U}(15,50)$ \\
  & b: $\mathcal{U}(0.0001, 0.1)$ & b: $\mathcal{U}(0.0001, 0.1)$  \\
  & $\tau_{out}$: $\mathcal{U}(15,25)$ & $\tau_{out}$: $\mathcal{U}(15,25)$ \\
loss aggregation & per timestep CE & per timestep CE \\
prediction mode & mean over sequence & mean over sequence \\
$t_0$ & 500(300) & 1023 \\
surrogate gradient & DG(SLAYER) & DG \\
grad.~clipping $m_\text{grad}$ & 1 & 1 \\
\bottomrule
\end{tabular*}
\end{table}
\subsection{Compute resources}
\label{app:compute_resources}
From the benchmark datasets, the Pathfinder-E experiments took the longest to execute, with $\approx$ 25 hours of execution time for 300 epochs on a single L40 GPU. The SHD experiments were the shortest, with $\approx$ 5 minutes of execution time for 300 epochs on an L40 GPU.  
\subsection{Code availability}
\label{app:code_availability}
The code to reproduce all experiments from this work is publicly available on GitHub under the CC BY-SA 4.0 license\footnote{\url{https://creativecommons.org/licenses/by-sa/4.0/}}  at \url{https://github.com/IMLTUGraz/HYPR}.

\subsection*{Acknowledgements}
This research was funded in whole or in part by the Austrian Science Fund (FWF) [10.55776/COE12] (R.L., M.B., OÖ), and by NSF EFRI grant \#2318152 (R.L., Y.B.). 
\clearpage
\newpage
\bibliographystyle{unsrtnat}
\bibliography{bibliography}

\end{document}